%% file: main.tex
\documentclass[manuscript,screen]{jair}

\setcopyright{none}
\makeatletter
\renewcommand\footnotetextcopyrightpermission[1]{}
\def\@mkbibcitation{}
\AtBeginDocument{%
  \fancyfoot{}%
  \fancypagestyle{firstpagestyle}{\fancyhf{}}%
}
\makeatother

\RequirePackage[
  datamodel=acmdatamodel,
  style=numeric-comp,
  backend=biber,
  natbib=true,
  giveninits=true,
  uniquename=init,
  sorting=none
]{biblatex}

\usepackage{booktabs}
\usepackage{multirow}
\usepackage{enumitem}
\usepackage{amsmath}
\usepackage{array}
\usepackage{graphicx}
\usepackage{url}
\usepackage{xspace}
\usepackage{tabularx}
\usepackage{threeparttable}
\usepackage{adjustbox}
\usepackage{pifont}
\usepackage{xcolor}
\usepackage{makecell}

\newcolumntype{Y}{>{\raggedright\arraybackslash}X}

\makeatletter
\authorsaddresses{}

\makeatother

\definecolor{emailgray}{gray}{0.1}

\begin{document}

\title[LLM Post-Training]{Large Language Model Post-Training: A Unified View of Off-Policy and On-Policy Learning}

\author{Shiwan Zhao}
\authornote{Corresponding authors: \nolinkurl{zhaosw@gmail.com, qinyong@nankai.edu.cn}.}
\affiliation{%
  \institution{Nankai University}
  \city{Tianjin}
  \country{China}
}

\author{Zhihu Wang}
\affiliation{%
  \institution{Huawei Technologies Ltd.}
  \city{Beijing}
  \country{China}
}

\author{Xuyang Zhao}

\author{Jiaming Zhou}
\affiliation{%
  \institution{Nankai University}
  \city{Tianjin}
  \country{China}
}

\author{Caiyue Xu}

\author{Chenfei Liu}
\affiliation{%
  \institution{Huawei Technologies Ltd.}
  \city{Beijing}
  \country{China}
}

\author{Liting Zhang}

\author{Yuhang Jia}

\author{Yanzhe Zhang}

\author{Hualong Yu}

\author{Zichen Xu}

\author{Qicheng Li}

\author{Yong Qin}
\authornotemark[1]
\affiliation{%
  \institution{Nankai University}
  \city{Tianjin}
  \country{China}
}

\renewcommand{\shortauthors}{Zhao et al.}

\begin{abstract}
\input{sections/abstract}
\end{abstract}


\maketitle

\input{sections/intro}
\input{sections/preliminaries}
\input{sections/unified_framework}
\input{sections/offpolicy_posttraining}
\input{sections/onpolicy_posttraining}
\input{sections/hybrid_comparative_analysis}
\input{sections/emerging_directions_and_open_problems}
\input{sections/conclusion}

\printbibliography

\end{document}

%% file: sections/abstract.tex
\noindent\textbf{Abstract—}Post-training has become a central phase in transforming broadly pretrained large language models (LLMs) into aligned, task-competent, and deployable systems. Recent progress spans supervised fine-tuning (SFT), preference optimization, reinforcement learning (RL), process supervision, verifier-guided methods, distillation, and increasingly elaborate multi-stage pipelines. Yet these methods are still often discussed in fragmented ways, organized by historical labels or objective families rather than by the behavioral bottlenecks they address.

This survey argues that LLM post-training is best understood as structured intervention on model behavior. We first organize the field by \emph{trajectory provenance}, which defines two primary learning regimes: \emph{off-policy learning}, which improves the model on externally supplied trajectories, and \emph{on-policy learning}, which improves it on learner-generated rollouts. We then interpret major methods through two recurring distribution-level roles---\emph{effective support expansion}, which makes useful behaviors more reliably reachable, and \emph{policy reshaping}, which improves behavior within already reachable regions---together with a complementary systems-level role, \emph{behavioral consolidation}, which preserves, transfers, and amortizes useful behavior across post-training stages and model transitions.

The resulting framework yields a unified reading of major post-training paradigms. Under this view, SFT may serve either support expansion or policy reshaping; offline preference optimization is usually off-policy reshaping, while online preference optimization brings preference-based supervision closer to learner-generated states. On-policy RL often improves behavior on learner-generated states, though stronger guidance can also make previously hard-to-reach reasoning paths effectively reachable. Distillation is often better understood as a consolidation mechanism rather than only as compression, and hybrid pipelines emerge as coordinated multi-stage compositions rather than ad hoc objective stacks.

Overall, the framework helps diagnose post-training bottlenecks and reason about stage composition, suggesting that progress in LLM post-training increasingly depends on coordinated systems design across regimes, roles, and stages rather than on any single dominant objective.

%% file: sections/intro.tex
\section{Introduction}
\label{sec:introduction}

In modern LLM systems, pretraining alone is no longer sufficient: post-training has become a central phase in turning broadly pretrained models into aligned, task-competent, and deployable systems. In practice, this phase is implemented through a diverse set of techniques, including supervised fine-tuning (SFT), preference optimization, reinforcement learning (RL), process supervision and verifier-guided methods, distillation, and increasingly elaborate multi-stage pipelines \citep{wei2022finetuned, rafailov2023dpo, ouyang2022rlhf, lightman2023lets, guan2024verifier, gu2024minillm, glm5team2026glm5}. These methods now shape instruction following, reasoning, safety alignment, controllability, and deployment efficiency. Recent system reports further make clear that post-training design has become a central source of capability differences among frontier systems \citep{grattafiori2024llama, team2025kimi, liu2025deepseek, glm5team2026glm5}.

Despite this rapid progress, the conceptual understanding of post-training remains fragmented. Much of the literature is still organized by historical labels or objective families: instruction tuning is discussed separately from preference optimization; reinforcement learning from human feedback (RLHF) and related alignment-oriented post-training methods are often discussed separately from reinforcement learning with verifiable rewards (RLVR) and related reasoning-oriented post-training methods; and distillation is often treated mainly as compression rather than as part of a broader post-training process \citep{zhang2023instruction_tuning_survey, liu2025surveydpo, wang2023aligning, zhang2025reasoningsurvey, xu2024kd_survey_llm}. This fragmentation obscures an increasingly important fact: many of these methods are not isolated alternatives, but different ways of intervening on the same underlying object---model behavior as induced by the trajectory distribution. As a result, comparisons are often made at the level of objective form while overlooking deeper behavioral questions: what kinds of behavior a method can introduce, what failures it can directly observe, and why strong systems so often combine multiple paradigms rather than relying on any single one.

These questions have become more pressing as post-training has shifted from relatively simple instruction-tuning pipelines toward explicitly multi-stage systems. Early instruction-tuned models were typically framed as supervised adaptation on curated instruction-response examples \citep{wei2022finetuned, sanh2022t0, wang2022supernatural, wang2023selfinstruct, taori2023alpaca, zhou2023lima, chung2024scaling}. More recent systems increasingly combine offline supervision with preference optimization, online reinforcement learning, process- or verifier-guided feedback, replay-enhanced policy optimization, and later-stage teacher-guided transfer \citep{rafailov2023dpo, ouyang2022rlhf, wang2024mathshepherd, li2025repo, deepseek2025r1, glm5team2026glm5}. The field therefore looks less like a menu of isolated objectives and more like a systems problem: how to establish useful behavior, refine it on learner-generated states, and preserve it across stage transitions.

\begin{figure*}[t]
    \centering
    \includegraphics[width=0.98\textwidth]{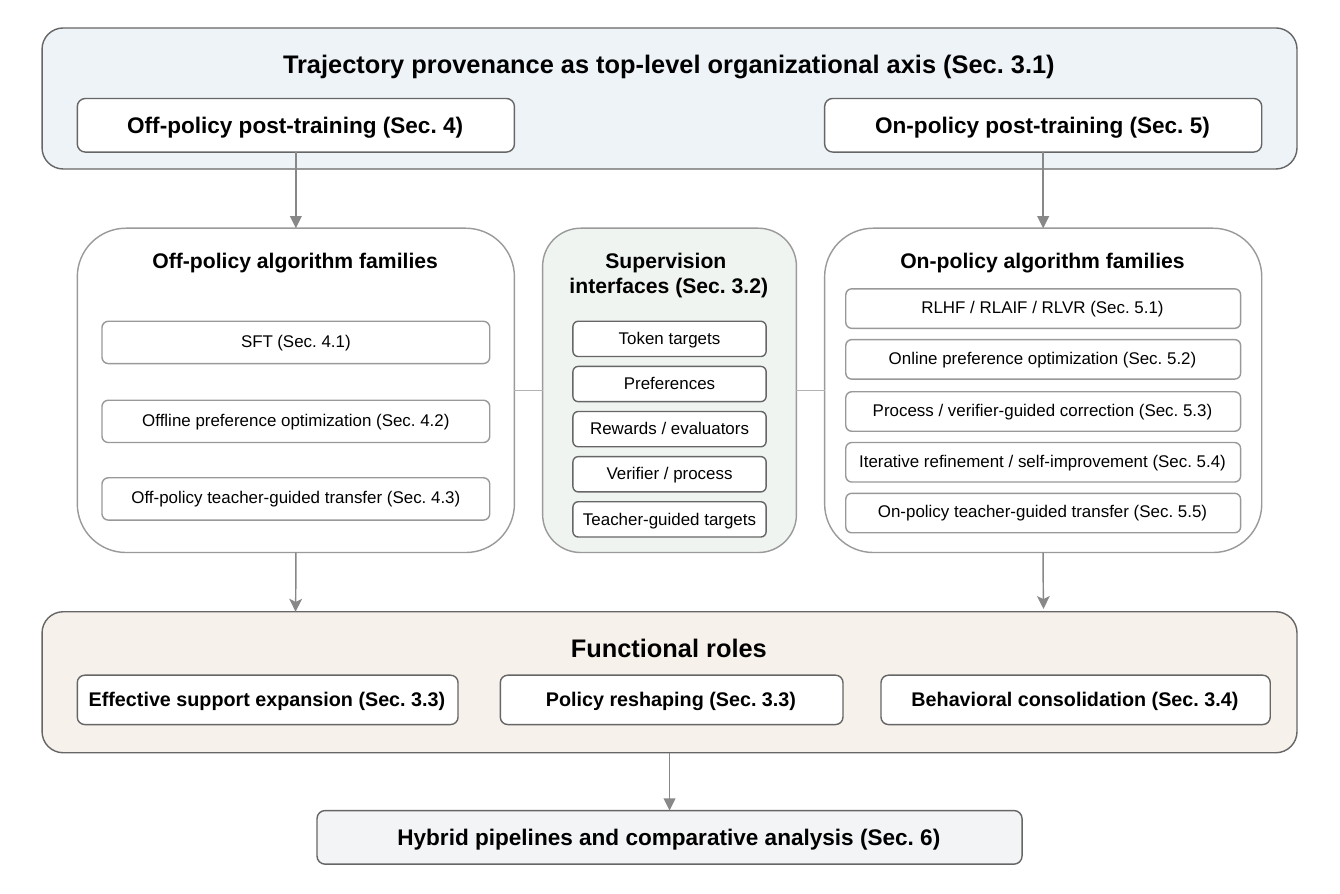}
    \caption{
    Unified overview of the survey's organizational and analytical framework. Trajectory provenance serves as the top-level organizational axis, which motivates the review of off-policy and on-policy post-training in Sections~\ref{sec:offpolicy} and~\ref{sec:onpolicy}. Within each regime, the method sections are organized by major algorithm families. Supervision interfaces are retained as a secondary descriptive layer for clarifying how training signals are attached within and across method families, whereas functional roles serve as the main explanatory lens for interpreting behavioral change. Hybrid pipelines are synthesized later in Section~\ref{sec:hybrid_comparative} as coordinated multi-stage compositions rather than as a third regime parallel to off-policy and on-policy learning. The figure is intended as both a reading map and a compact preview of the framework formalized in Section~\ref{sec:unified_framework}; it does not imply fixed one-to-one mappings between regime, interface, and role.
    }
    \label{fig:intro_outline}
\end{figure*}

This survey argues that a productive way to organize this landscape is to begin by distinguishing between two learning regimes defined by \emph{trajectory provenance}: \emph{off-policy learning}, which improves the model on externally supplied trajectories, and \emph{on-policy learning}, which improves it on learner-generated rollouts. This top-level distinction organizes the survey’s main method sections. Within each regime-specific section, methods are further grouped by major algorithm families for readability and comparison. \emph{Supervision interfaces} are retained as a secondary descriptive layer, while \emph{functional roles} provide the paper's main explanatory lens. Formal definitions are developed in Section~\ref{sec:unified_framework}; Figure~\ref{fig:intro_outline} provides a compact overview of this organization.

\paragraph{Scope of this survey}
This survey focuses on post-training methods for LLMs following general pretraining, especially those aimed at improving instruction following, alignment, reasoning, and other deployment-relevant capabilities. We cover supervised fine-tuning, preference-based learning, RLHF- and RLVR-style methods, process supervision and verifier-guided methods, distillation and teacher-guided transfer, and hybrid post-training systems. Where useful, we also draw on work in deployment-coupled settings such as red teaming, retrieval-augmented reasoning, tool use, and agentic system reports when they provide evidence about post-training design or the transfer of post-trained behavior \citep{perez2022redteaming, zhang2025reasonrag, schick2023toolformer, yao2022react, glm5team2026glm5}. We do not attempt a full survey of pretraining, retrieval augmentation, inference-time scaling, or tool use as independent topics, although these sometimes enter the discussion when they materially affect post-training design.

\paragraph{Literature coverage and curation note}
This survey is a representative, concept-driven survey rather than an exhaustive systematic review. We prioritize papers and system reports that are most useful for understanding post-training design, including foundational methods, widely used objectives, important reasoning and alignment variants, transfer-oriented methods, and frontier multi-stage pipelines. Archival papers provide the backbone of the survey's conceptual claims, while recent system reports and selected preprints are used more selectively to illustrate emerging design patterns in practice. Throughout this survey, classifications such as ``off-policy'' versus ``on-policy'' or ``support expansion'' versus ``policy reshaping'' should be understood as dominant analytical assignments within the present framework, not as immutable labels.

\begin{table*}[t]
\raggedright
\caption{Qualitative positioning of representative prior surveys under the organizational lens adopted in this paper. `Off/On axis' asks whether off-policy versus on-policy learning is used as a top-level organizing principle; `Functional roles' asks whether methods are analyzed by their dominant behavioral role rather than mainly by method family; `Distill.\ transfer' asks whether distillation is treated as cross-stage transfer rather than only as compression; and `Multi-stage synthesis' asks whether hybrid and related multi-stage compositions are synthesized explicitly. The comparison reflects the specific organizational lens of this work rather than a general assessment of survey quality, coverage, or depth. The symbols indicate recurring organizational emphasis: $\bullet$ denotes a recurring primary organizing emphasis, $\circ$ denotes secondary or implicit treatment, and $-$ denotes that the dimension is not foregrounded as a recurring organizing principle.}
\label{tab:prior-survey-comparison}
\small
\renewcommand{\arraystretch}{1.14}
\setlength{\tabcolsep}{4pt}
\begin{tabularx}{\textwidth}{>{\raggedright\arraybackslash}p{3.0cm} c c c c >{\raggedright\arraybackslash}X}
\toprule
\textbf{Survey cluster}
& \makecell[c]{\textbf{Off/On}\\\textbf{axis}}
& \makecell[c]{\textbf{Functional}\\\textbf{roles}}
& \makecell[c]{\textbf{Distill.}\\\textbf{transfer}}
& \makecell[c]{\textbf{Multi-stage}\\\textbf{synthesis}}
& \textbf{Main emphasis} \\
\midrule

Instruction tuning
{\scriptsize \citep{zhang2023instruction_tuning_survey}}
& $-$ & $-$ & $-$ & $-$
& Instruction tuning / supervised fine-tuning, instruction following, dataset construction, and applications across domains \\

Preference optimization
{\scriptsize \citep{liu2025surveydpo}}
& $-$ & $-$ & $-$ & $-$
& Direct preference optimization, preference optimization variants, benchmarking, applications, and open challenges \\

RLHF / alignment
{\scriptsize \citep{casper2023open, wang2023aligning}}
& $-$ & $-$ & $-$ & $\circ$
& LLM alignment methods, RLHF, evaluation, and RLHF limitations / open problems \\

Process / verifier
{\scriptsize \citep{zheng2025prmSurvey, guan2024verifier}}
& $-$ & $\circ$ & $-$ & $\circ$
& Process reward modeling, process supervision, verifier engineering, and verifier-guided post-training \\

Distillation
{\scriptsize \citep{xu2024kd_survey_llm}}
& $-$ & $\circ$ & $\bullet$ & $-$
& Teacher--student transfer, compression, self-improvement, and knowledge distillation methods \\

General post-training
{\scriptsize \citep{tie2025posttraining_survey}}
& $-$ & $-$ & $-$ & $-$
& Broad post-training taxonomy spanning fine-tuning, alignment, reasoning, efficiency, and integration / adaptation \\

Post-training scaling
{\scriptsize \citep{lai2025posttrainingscaling}}
& $-$ & $-$ & $-$ & $-$
& Post-training scaling via SFT, RLxF, and test-time compute, emphasizing scalable methods and scaling motivation \\

Reasoning post-training
{\scriptsize \citep{kumar2025reasoningposttraining, zhang2025reasoningsurvey}}
& $-$ & $-$ & $-$ & $-$
& Reasoning-oriented post-training, including RL for large reasoning models and inference-time scaling \\

\midrule
\textbf{This survey}
& \textbf{$\bullet$} & \textbf{$\bullet$} & \textbf{$\bullet$} & \textbf{$\bullet$}
& \textbf{A unified view of post-training via trajectory provenance, functional roles, and multi-stage cross-regime composition} \\

\bottomrule
\end{tabularx}
\end{table*}

\paragraph{What this survey aims to contribute}
Beyond offering a cleaner taxonomy, this survey provides a unified framework for organizing, comparing, and interpreting LLM post-training. Rather than treating major paradigms as isolated objective families, it analyzes them through a common set of questions about where learning occurs, how supervision is attached, and what kind of behavioral change or systems role a stage primarily serves. In practical terms, this lens is meant to help diagnose when useful behavior is best imported through external supervision, when it should instead be corrected on learner-generated states, and when it must be preserved or transferred across stage boundaries.

This survey makes four main contributions.

\textbf{First}, it proposes trajectory provenance as a unified top-level organizational axis for LLM post-training, using the distinction between off-policy learning and on-policy learning to structure the landscape by where optimization actually occurs.

\textbf{Second}, it retains supervision interfaces as a secondary descriptive layer, so that token targets, preferences, reward-derived signals, verifier- or process-based feedback, and teacher-guided targets can be compared within a shared vocabulary across post-training regimes.

\textbf{Third}, it introduces a functional-role analysis centered on effective support expansion, policy reshaping, and behavioral consolidation, thereby providing a common explanatory vocabulary for characterizing the dominant behavioral effect or systems role of a post-training stage.

\textbf{Fourth}, it uses this layered framework to compare major paradigms within a single analytical language, explain why hybrid pipelines emerge naturally as coordinated multi-stage systems, and surface open problems concerning learner-state correction, support attrition, consolidation, and stage composition.

\paragraph{Relation to prior surveys and reviews}
Table~\ref{tab:prior-survey-comparison} situates this survey relative to prior reviews of major strands of the post-training literature, including instruction tuning, preference optimization, RLHF and alignment, process reward modeling and verifier-oriented methods, distillation, broader taxonomic overviews of post-training, post-training scaling, and reasoning-oriented post-training \citep{zhang2023instruction_tuning_survey, liu2025surveydpo, casper2023open, wang2023aligning, xu2024kd_survey_llm, zheng2025prmSurvey, tie2025posttraining_survey, guan2024verifier, lai2025posttrainingscaling, kumar2025reasoningposttraining, zhang2025reasoningsurvey}. The table is intended to highlight differences in recurring organizational emphasis under the present framework rather than to score coverage, depth, or quality. Several prior surveys discuss some of these dimensions substantially even when they do not foreground them as primary organizing principles; this broader pattern also appears in related reviews of preference tuning and RL--LLM interaction \citep{winata2025preference, pternea2024rltree}.

The main difference from prior surveys is therefore one of organization rather than judgment. Many prior reviews are structured around particular method families, application domains, or narrower slices of the alignment and reasoning landscape. By contrast, the present survey is organized around trajectory provenance as a top-level axis. It uses functional-role analysis to compare both behavioral change within stages and preservation across stages, treats distillation as cross-stage transfer rather than only as compression, and explicitly synthesizes multi-stage post-training compositions. Within this framework, supervision interfaces are treated as a secondary descriptive layer rather than as the highest-level organizing principle.

\paragraph{Organization of the survey}
Section~\ref{sec:preliminaries} introduces the minimum formal and conceptual background. Section~\ref{sec:unified_framework} then formalizes the survey's unified framework. Sections~\ref{sec:offpolicy} and~\ref{sec:onpolicy} review off-policy and on-policy post-training, respectively. Section~\ref{sec:hybrid_comparative} synthesizes hybrid pipelines and compares major paradigms through the unified lens. Section~\ref{sec:emerging_open} discusses emerging directions and unresolved open problems, and Section~\ref{sec:conclusion} concludes.

%% file: sections/preliminaries.tex
\section{Preliminaries and Problem Formulation}
\label{sec:preliminaries}

This section introduces the minimum conceptual and formal background needed for the rest of the survey. Its purpose is not to present the paper's full interpretive framework, but to establish a common language for describing LLM post-training. We first view an autoregressive language model as a policy over token trajectories, then distinguish off-policy from on-policy learning at a basic level, and finally formulate post-training as the problem of improving deployment-relevant behavior after general pretraining. The more explicitly interpretive concepts used throughout this survey---including effective support expansion, policy reshaping, and behavioral consolidation---are developed in Section~\ref{sec:unified_framework} rather than assumed here.

\subsection{Language Models as Policies over Trajectories}
\label{subsec:lm_as_policy}

An autoregressive language model defines a conditional distribution over output token sequences given an input prompt. Let $x$ denote an input prompt and let $y = (y_1, y_2, \dots, y_T)$ denote an output sequence. The model defines
\begin{equation}
\pi_\theta(y \mid x) = \prod_{t=1}^{T} \pi_\theta(y_t \mid x, y_{<t}),
\end{equation}
where $\theta$ denotes the model parameters and $y_{<t}$ denotes the token sequence generated before step $t$.

This factorization naturally admits a sequential decision-making view. At each step, the current prefix $(x, y_{<t})$ can be interpreted as a state, the next token $y_t$ as an action, and the full output sequence as a trajectory. The familiar prefix-based view of language modeling is therefore not separate from a sequential state-action view; the prefix simply plays the role of state in the autoregressive process. This perspective has become increasingly common in work on RLHF, preference optimization, process supervision, and reasoning-oriented RL, where policies and trajectories---and often learner-generated rollouts---are explicit objects of analysis \citep{ouyang2022rlhf, rafailov2023dpo, lightman2023lets, shao2024deepseekmath, deepseek2025r1}. It is also naturally related to standard policy-based views in reinforcement learning \citep{schulman2015trpo, schulman2017ppo, sutton1998reinforcement}.

The point is not that language modeling simply reduces to classical reinforcement learning. Rather, the policy-over-trajectories view provides a common abstraction for comparing post-training methods that intervene on sequence generation in different ways. Some optimize token likelihood under external demonstrations; others optimize preferences or rewards defined over sampled responses; still others transfer behavior from stronger teachers. In all cases, the model induces a trajectory distribution that shapes which behaviors are likely to occur at deployment time, which states are likely to be visited along the way, and which local decisions become reliably available once those states are reached.

\subsection{Off-Policy and On-Policy Learning}
\label{subsec:basic_offpolicy_onpolicy}

A foundational distinction in this survey is whether learning is driven primarily by externally supplied trajectories or by learner-generated trajectories sampled from the current model.

\paragraph{Off-policy learning}
In off-policy learning, the model is updated using trajectories that are not sampled from its current policy. These may include instruction-response pairs, expert demonstrations, paired or ranked preference data, verified solutions, synthetic teacher outputs, or replayed traces from earlier policies. Supervised fine-tuning and many standard preference-optimization methods in their common offline form are representative examples \citep{wei2022finetuned, wang2023selfinstruct, zhou2023lima, chung2024scaling, rafailov2023dpo, hong2024orpo, xu2024cpo, ethayarajh2024kto, meng2024simpo}.

\paragraph{On-policy learning}
In on-policy learning, the model is updated using trajectories sampled from its current policy and feedback computed on those newly generated rollouts. Such feedback may be derived from rewards, online preference judgments, constitutional rules, tests, or verifiers. RLHF, constitutional RL, and RL with verifiable rewards are standard examples when optimization is driven by freshly sampled model trajectories \citep{ziegler2019lm_preferences, stiennon2020summarize, ouyang2022rlhf, bai2022hh_rlhf, bai2022constitutional, shao2024deepseekmath, deepseek2025r1}.

At a basic level, this distinction matters because it determines which states the learner is trained on and how supervision is exposed to it. Off-policy learning can expose the model to behaviors that it would not yet reliably generate on its own, whereas on-policy learning can directly improve behavior on the states the model actually visits under its own rollout dynamics. This provenance-centered distinction is related to, but not identical with, the broader offline and online reinforcement learning literature \citep{levine2020offline, prudencio2023survey}. In the RL literature, \emph{offline} versus \emph{online} usually tracks whether optimization proceeds from a fixed dataset or continues to interleave with fresh environment interaction. Here, by contrast, \emph{off-policy} versus \emph{on-policy} tracks where the trajectories used to update the learner come from: externally supplied traces or rollouts generated by the current learner itself. We invoke the offline/online distinction only as background context rather than as a direct template for LLM post-training taxonomy.

\subsection{Occupancy, Prefix Reachability, and Action Availability}
\label{subsec:saom_prelim}

As preparatory background, it is useful to borrow an occupancy-style view of state visitation and state-conditional action choice from reinforcement learning \citep{puterman1994markov, altman1999cmdp}. We use it here as a compact analytic vocabulary for decomposing autoregressive behavior into two coupled components: which prefixes the model tends to visit, and which next-token decisions it tends to make once those prefixes are reached. Some recent language-modeling work has similarly adopted occupancy-style views to analyze preference optimization, multi-step alignment, and multi-turn interaction settings \citep{shi2024direct, wu2025multistep}. The purpose of this vocabulary is simple: to separate whether useful prefixes are reached from whether useful next actions receive sufficient probability once those prefixes are reached.

In the LLM setting, a state at step $t$ is the prompt-prefix pair
\[
s_t = (x, y_{<t}),
\]
and the action is the next token
\[
a_t = y_t.
\]
Under a prompt distribution $\mathcal{D}_x$, we define a normalized state-action occupancy measure of policy $\pi_\theta$ as
\begin{equation}
d^{\pi_\theta}_{\mathcal{D}_x}(s,a)
=
\mathbb E_{x\sim \mathcal{D}_x,\; y\sim \pi_\theta(\cdot\mid x)}
\left[
\sum_{t=1}^{T(y)}\eta_t(y)\,\mathbf 1\{s_t=s,\; a_t=a\}
\right],
\label{eq:saom}
\end{equation}
where $\eta_t(y) \ge 0$ and $\sum_{t=1}^{T(y)} \eta_t(y)=1$ for each sampled sequence $y$. Here, $s_t$ denotes the prefix state observed at position $t$, and $a_t$ denotes the token selected at that position. Intuitively, $d^{\pi_\theta}_{\mathcal{D}_x}(s,a)$ aggregates normalized visitation mass of $(s_t,a_t)$ across trajectories generated under the prompt distribution and policy. For a sampled sequence of length $T(y)$, the simplest choice is uniform weighting, $\eta_t(y)=1/T(y)$, which treats all positions equally, although other weightings are also possible when certain positions should contribute more strongly.

A corresponding state occupancy measure is
\begin{equation}
d^{\pi_\theta}_{\mathcal{D}_x}(s)
=
\sum_a d^{\pi_\theta}_{\mathcal{D}_x}(s,a),
\label{eq:state_visitation}
\end{equation}
which captures how much normalized visitation mass the policy assigns to state $s$ under the prompt distribution and rollout dynamics.

This decomposition is useful because post-training is not only about assigning higher probability to desirable final responses. It is also about which intermediate prefixes become practically reachable and which locally important actions become available along the way. In autoregressive models, the next state is formed by appending the chosen token to the current prefix, so local action changes can recursively alter future state visitation. This makes \emph{prefix reachability} and \emph{state-conditional action availability} natural coupled objects of analysis, especially in long-horizon reasoning where small local changes can redirect generation toward very different downstream trajectories.

For this reason, it is useful to distinguish between two notions:
\begin{itemize}[leftmargin=2em]
    \item \textbf{Prefix/state reachability:} whether useful states are visited with non-negligible mass under the policy, as reflected by $d^{\pi_\theta}_{\mathcal{D}_x}(s)$;
    \item \textbf{State-conditional action availability:} whether useful actions receive sufficient probability once those states are reached, as reflected locally by $\pi_\theta(a\mid s)$ and jointly by $d^{\pi_\theta}_{\mathcal{D}_x}(s,a)$.
\end{itemize}

The same perspective also clarifies an important source of mismatch between training and deployment. Under teacher-forced or otherwise externally curated supervision, optimization is often carried out on prefixes drawn from an external data distribution, whereas at deployment the model must continue from prefixes induced by its own rollout dynamics. In occupancy-style terms, off-policy supervision may act primarily on externally supplied state-action visitation patterns, whereas on-policy learning acts on visitation induced by the current policy. When these differ substantially, supervision may fail to directly correct errors that arise only on learner-generated states. Classical sequence-learning work studied related mismatch effects through ideas such as scheduled sampling, DAgger, and professor forcing \citep{bengio2015scheduled, ross2011dagger, lamb2016professor}. In LLM post-training, the same structural issue reappears whenever methods differ in whether they supervise externally supplied trajectories or learner-induced rollouts. These distinctions will later support the survey's separation between support expansion and policy reshaping.

\subsection{A Generic View of Post-Training Objectives}
\label{subsec:generic_posttraining_view}

At a descriptive level, many post-training methods can be expressed through an optimization template of the form
\begin{equation}
\max_\theta \; \mathbb E_{(x,y,z)\sim \mathcal D} \big[ \mathcal J(\theta; x,y,z) \big],
\label{eq:generic_pt_obj}
\end{equation}
where $\mathcal D$ denotes a source of supervision-bearing training tuples, $y$ denotes an output, trajectory, or rollout trace, and $z$ represents additional signals such as preference labels, rewards, verifier outcomes, critiques, or teacher guidance.

This abstraction is intentionally broad. It is useful for comparing where trajectories come from and how supervision is attached to them, but it is not yet an explanation of what kind of behavioral change a method induces. In off-policy settings, $\mathcal D$ may denote an external dataset of demonstrations, preferences, or verified traces; in on-policy settings, it may include tuples generated from the current learner's rollouts together with associated feedback signals. The explanatory step---distinguishing support expansion, policy reshaping, and behavioral consolidation---is deferred to Section~\ref{sec:unified_framework}.

\subsection{Problem Formulation: What Post-Training Must Achieve}
\label{subsec:problem_formulation}

With this descriptive template in place, we can state the post-training problem more directly. After large-scale pretraining, a language model typically contains broad statistical knowledge and substantial general capability, but it is not yet optimized for downstream requirements such as instruction following, safe interaction, faithful reasoning, controllability, or deployment efficiency. Post-training addresses this gap.

At a high level, the post-training problem can be formulated as follows: given a pretrained policy $\pi_{\theta_0}$ and a set of supervision sources---for example demonstrations, preferences, rewards, critiques, verifiers, or teacher models---learn an updated policy $\pi_\theta$ that behaves better under deployment-relevant prompts and constraints. What counts as ``better'' depends on the setting: it may involve higher preference quality, stronger reasoning reliability, lower harmfulness, better verifier success, greater deployment efficiency, or some combination of these objectives.

Crucially, these improvements do not all arise in the same way. Some require exposing the learner to trajectories it would not yet reliably reach on its own. Others require improving decisions on the learner's own rollout states. Still others require preserving useful behavior across stages, model transitions, or deployment settings. Post-training therefore often has to solve more than one kind of problem at once rather than optimize a single objective in isolation. Section~\ref{sec:unified_framework} turns these recurring demands into the survey's unified framework.

%% file: sections/unified_framework.tex
\section{A Unified Framework for LLM Post-Training}
\label{sec:unified_framework}

The previous section introduced the minimum vocabulary needed for this survey: an autoregressive language model can be viewed as a policy over trajectories; training may act on trajectories that differ from those induced by the model's own rollout dynamics at deployment; and deployment quality depends jointly on which states are reached and which local actions are selected there. We now formalize the layered framework previewed in the introduction and summarized in Figure~\ref{fig:intro_outline}.

Our main claim is that modern LLM post-training is most usefully understood as structured intervention on model behavior. The framework separates three analytical questions: where learning occurs, how supervision is attached, and what kind of behavioral change or systems role a stage primarily serves. The first question provides the survey's top-level organizational axis. The latter two clarify comparison and explanation within and across regimes.

\subsection{Trajectory Provenance as the Top-Level Organizing Axis}
\label{subsec:provenance_axis}

We begin with trajectory provenance because the most consequential distinction in post-training is often not the algebraic form of the loss, but the source of the trajectories on which learning occurs. In modern LLM systems, two top-level regimes dominate:
\begin{itemize}[leftmargin=2em]
    \item \textbf{Off-policy learning:} updates are driven primarily by externally supplied trajectories, such as demonstrations, preference pairs, verified traces, synthetic teacher outputs, or replayed traces relative to the current learner;
    \item \textbf{On-policy learning:} updates are driven primarily by rollouts sampled from the current policy and then improved using feedback derived from rewards, online preference judgments, constitutional rules, tests, or verifiers.
\end{itemize}

We adopt this distinction as the highest-level taxonomy because it most directly determines which states the learner is trained on, which failures are directly exposed during optimization, and which behavior regions can be improved directly versus only after additional exploration. Off-policy methods can expose the learner to useful trajectories that it would not yet reliably generate on its own. On-policy methods instead optimize on the states the learner actually visits under its own rollout dynamics, which becomes especially important when errors appear only after self-generated branching or compound over multi-step interaction.

Other distinctions remain important, but they are better treated as second-order differences. Methods may differ in objective family, feedback source, or supervision granularity; these differences affect how learning proceeds, but they are downstream of the more basic question of where the learner is optimized. For this reason, objective form is not the top-level organizing principle of this survey.

\subsection{Supervision Interfaces as a Secondary Descriptive Layer}
\label{subsec:supervision_interface_layer}

Within either provenance regime, learning can still proceed through very different supervision interfaces. By this term, we mean the form in which learning signals are attached to a given set of states or trajectories. Important examples include:
\begin{itemize}[leftmargin=2em]
    \item token targets, as in standard supervised fine-tuning;
    \item pairwise or ranked preferences, as in DPO-style or related preference-learning methods;
    \item reward-derived signals, as in RLHF or RLVR;
    \item verifier- or process-based feedback, as in verifier-guided methods or process supervision;
    \item teacher-guided targets, such as teacher logits, teacher-generated traces, teacher-filtered outputs, or other teacher-derived supervision in distillation-style transfer.
\end{itemize}

This layer matters because similar trajectory sources can still induce different learning dynamics, and provenance alone does not fully determine how supervision is attached once the training states have been fixed. Standard SFT and off-policy distillation, for example, may both operate on externally supplied traces, yet one uses fixed token targets whereas the other uses teacher-guided targets, outputs, or trajectories \citep{hinton2015distilling, snell2022context, huang2022icl_distill, hsieh2023distilling, shridhar2023distilling}. Likewise, on-policy RL and on-policy teacher-guided transfer may both act on learner-generated rollouts, yet they differ in whether improvement is driven primarily by scalar evaluative signals or by teacher-derived supervision on those visited states \citep{gu2024minillm, agarwal2024opd, ye2025blackbox_opd, ye2026opcd, zhao2026selfDistilledReasoner, yang2026gopd, sang2026crisp}. In this sense, supervision interfaces are not a competing top-level taxonomy, but a secondary descriptive layer that helps clarify important within-regime differences.

This distinction is especially useful for locating distillation. We do not treat distillation as a third trajectory regime parallel to off-policy and on-policy learning. Distillation is better understood as a teacher-guided transfer pattern that can be instantiated in either regime: when teacher-derived targets are attached to externally supplied traces, it is off-policy; when they are attached to learner-generated rollouts, it is on-policy. Its distinguishing feature is therefore not a separate source of trajectories, but teacher-mediated target construction within the supervision-interface layer.

\subsection{Distribution-Level Functional Roles}
\label{subsec:functional_roles}

Trajectory provenance and supervision interface specify where training acts and how learning signals are delivered, but they do not yet characterize the dominant kind of behavioral change that training induces. To address that question, we now introduce distribution-level functional roles.

Building on the occupancy-style preliminaries in Section~\ref{subsec:saom_prelim}, we define an effective-support notion on the state-action space under a fixed task distribution and use it to distinguish effective support expansion from policy reshaping. Trajectories remain the natural level for interpreting end-to-end behavior, but the primary formalization here is on realized state-action occupancy because it provides a more precise bridge between local optimization and sequential consequences.

\subsubsection{Effective Support on the State-Action Space}
\label{subsubsec:effective_support_definition}

In autoregressive language models, exact probabilistic support is usually too broad to be analytically useful: many tokens and many finite continuations have nonzero probability. For post-training analysis, the more relevant question is not whether a behavior is theoretically possible, but whether it is realized with non-negligible probability mass under realistic rollout dynamics.

For this reason, we work with an effective-support notion defined on the normalized state-action occupancy measure from Equation~\ref{eq:saom}. For a threshold $\epsilon > 0$, define
\begin{equation}
\mathcal E_\epsilon(\pi; \mathcal D_x) = \{(s,a): d^{\pi}_{\mathcal D_x}(s,a) \ge \epsilon\}.
\label{eq:effective_support}
\end{equation}
Intuitively, $\mathcal E_\epsilon(\pi; \mathcal D_x)$ collects the state-action patterns that are behaviorally realized with non-negligible mass under policy $\pi$ and prompt distribution $\mathcal D_x$.

This is an operational rather than strictly measure-theoretic notion. Recent work has begun to operationalize reduced empirical proxies for support at the answer/completion level under finite sampling \citep{wu2025support}. Although such proxies do not fully instantiate state-action support, they suggest that support-based analyses can admit measurable approximations in constrained settings. In this survey, effective support is defined on the state-action space because deployment-relevant behavior depends jointly on reaching a state and taking the right action there. It is especially informative in settings such as reasoning, verifier-guided correction, or guided multi-step search, where intermediate learner states materially affect final outcomes.

\subsubsection{Effective Support Expansion}
\label{subsubsec:support_expansion}

With Equation~\ref{eq:effective_support} in place, we say that effective support expansion occurs when training causes previously negligible state-action regions to enter effective support:
\begin{equation}
\mathcal E_\epsilon(\pi_1; \mathcal D_x) \setminus \mathcal E_\epsilon(\pi_0; \mathcal D_x) \neq \varnothing.
\label{eq:support_expansion}
\end{equation}
That is, some state-action patterns that were previously behaviorally negligible under $\pi_0$ become meaningfully occupied under $\pi_1$. This condition identifies the presence of expansion; whether a stage is expansion-dominant is a later interpretive judgment about which bottleneck the stage primarily resolves.

Unless the context requires emphasis or added precision, we use \emph{support expansion} as shorthand for \emph{effective support expansion} in the remainder of this paper.

At the autoregressive level, expansion can occur when training increases effective support on state-action pairs---for example, by making local actions reliably selectable in states where they were previously ineffective---thereby opening downstream behavioral branches that were effectively unavailable before. At the trajectory level, the same phenomenon appears as making previously negligible but useful trajectories more practically reachable and completable. This is why off-policy methods are often well positioned to induce expansion: they can supervise trajectories and local decisions that the learner would not yet reliably realize on its own.

\subsubsection{Policy Reshaping}
\label{subsubsec:policy_reshaping}

Policy reshaping refers to a different kind of distributional intervention from support expansion. Here, the relevant states and actions are already effectively available, but the policy allocates probability mass poorly within that already reachable region. Training then improves behavior mainly by redistributing probability mass within existing effective support rather than by making genuinely new state-action regions effective.

Let
\begin{equation}
\mathcal C_\epsilon(\pi_0,\pi_1; \mathcal D_x) = \mathcal E_\epsilon(\pi_0; \mathcal D_x) \cap \mathcal E_\epsilon(\pi_1; \mathcal D_x)
\label{eq:shared_support}
\end{equation}
denote the shared effective support before and after training. We say that training is reshaping-dominant when its main behavioral effect is explained by probability redistribution within $\mathcal C_\epsilon(\pi_0,\pi_1; \mathcal D_x)$ rather than by activation of state-action pairs outside this shared support.

At the trajectory level, policy reshaping makes better trajectories more probable and poorer ones less so among behaviors that were already effectively reachable. At the finer autoregressive level, it typically reallocates state-conditional action probabilities on prefixes the learner already reaches with meaningful frequency, primarily among action choices that already lie within shared effective support. In this sense, reshaping is best understood as redistribution within shared effective state-action support, not as activation of genuinely new effective regions.

This notion is broad enough to include improved ranking among candidate responses, better alignment to preference signals, stronger style calibration, better local branch choice in reasoning, or more reliable recovery from intermediate mistakes.

\subsubsection{Support Attrition as a Diagnostic Complement}
\label{subsubsec:support_attrition}

It is also useful to track the loss of previously effective behavior. Define the attrited region as
\begin{equation}
\mathcal A_\epsilon(\pi_0,\pi_1; \mathcal D_x) = \mathcal E_\epsilon(\pi_0; \mathcal D_x) \setminus \mathcal E_\epsilon(\pi_1; \mathcal D_x).
\label{eq:attrition}
\end{equation}
This set contains state-action patterns that were effectively realized before training but no longer remain in effective support afterward.

Support attrition is not a third positive distribution-level role alongside support expansion and policy reshaping. Rather, it is a complementary diagnostic notion for analyzing forgetting, regression, or stage-to-stage capability loss. Catastrophic forgetting \citep{mccloskey1989catastrophic, kirkpatrick2017overcoming}, for example, may arise not only from harmful reshaping within retained regions, but also from the outright disappearance of previously effective behavior.

\subsubsection{Boundary Cases and Role Overlap}
\label{subsubsec:role_overlap}

The distinction between support expansion and policy reshaping is analytically useful, but it should not be interpreted as perfectly rigid. The same method family may play different dominant roles in different settings: SFT may function as expansion when it introduces behavior the model would not otherwise realize, but as reshaping when it mainly calibrates style, alignment, or instruction-conditioned response selection. On-policy RL is often reshaping-dominant because its updates are concentrated on learner-visited states, yet under sufficiently strong exploration or structured intermediate feedback it may also contribute to support expansion.

Local action-level change can also induce trajectory-level expansion: if training makes previously negligible but crucial local actions reliably selectable, downstream trajectory regions may become effectively reachable as a consequence. This is one reason effective support is defined on the state-action space rather than on states alone.

Role overlap is common. A method may simultaneously activate previously negligible state-action regions, redistribute mass within retained regions, and cause attrition elsewhere. The value of the framework therefore lies in identifying the dominant intervention role in context rather than in pretending that mixed cases do not exist.

\subsection{Behavioral Consolidation as a Systems-Level Role}
\label{subsec:behavioral_consolidation}

The roles above characterize how behavior changes within a training stage. Modern post-training systems, however, are often multi-stage. This introduces a different question: even if a stage induces useful behavior, will that behavior survive handoff to a later model, a cheaper deployment policy, or a simplified runtime setting? To capture this, we introduce behavioral consolidation as a systems-level role. This asymmetry is intentional: support expansion and policy reshaping describe distribution-level changes within or across learner-relevant regions, whereas behavioral consolidation captures the systems problem of preserving, transferring, stabilizing, or amortizing useful behavior across stage boundaries, model transitions, or deployment regimes.

Concretely, behavioral consolidation concerns whether useful induced behavior survives handoff across stages, models, or deployment settings, and in what form it can be preserved, transferred, amortized, or compressed. This includes, for example, distilling a strong but expensive policy into a smaller model, transferring behavior learned under scaffolded interaction into a standalone model, or compressing multi-turn agentic behavior into a deployable policy. In this sense, behavioral consolidation is not a third distributional role alongside support expansion and policy reshaping, but a systems-level question about whether improvements remain available after regime changes.

Typical forms include:
\begin{itemize}[leftmargin=2em]
    \item \textbf{Retention across stages:} preserving gains when moving from one optimization regime to another;
    \item \textbf{Transfer:} carrying behaviors discovered by a teacher, search process, verifier-guided pipeline, or stronger policy into a new student or successor policy;
    \item \textbf{Amortization:} compiling expensive runtime behaviors such as reflection, search, or multi-pass verification into a cheaper direct policy;
    \item \textbf{Compression:} distilling useful behavior into a smaller or lower-cost deployment model without excessive loss.
\end{itemize}

Consolidation is closely related to---but distinct from---support attrition. Attrition describes what is lost at the distribution level, namely state-action patterns that drop out of effective support after training. Consolidation instead describes a systems objective: ensuring that useful behavior can be preserved, transferred, compressed, or internalized into a more deployable model rather than being erased, diluted, or remaining dependent on auxiliary context or other costly external support \citep{snell2022context, gu2024minillm}. This perspective is especially useful for understanding distillation. Although distillation can in principle contribute to either expansion or reshaping depending on how teacher guidance is instantiated, in many modern LLM pipelines it is best understood as consolidation-dominant, because its central role is often to carry forward behaviors that were first induced elsewhere into a form that can be reused, deployed, or further transferred \citep{gu2024minillm, hsieh2023distilling, shridhar2023distilling, luo2026agentark}. Behavioral consolidation, however, is broader than distillation alone and may also be supported by mechanisms such as replay-based retention or model-level synthesis \citep{wang2024inscl, he2025mergebench}. Still, distillation remains the clearest and most prominent instantiation of this role in current LLM post-training practice and is therefore the main focus here.

\subsection{Operational Heuristics for Dominant Functional Roles}
\label{subsec:operational_heuristics}

The categories introduced above are not rigid labels, but dominant functional tendencies. In applying this framework, the first step is to fix the relevant task or prompt distribution under which behavior is being evaluated, since effective support and functional change are always relative to a distribution of interest. The second step is to choose an observational granularity: some questions are best examined at the trajectory level, while others require the finer state-action view that distinguishes prefix reachability from local decision quality. The third step is to identify the primary bottleneck. When a method both activates new behavior and improves behavior inside previously reachable regions, the most informative question is usually which bottleneck has actually been removed: lack of reachable useful behavior, poor choice within already reachable behavior, or failure to preserve useful behavior across stages.

Table~\ref{tab:operational_role_proxies} summarizes practical cues for assigning dominant roles in context. These cues are heuristic rather than decisive: mixed cases are common, and multi-stage systems may involve different dominant roles at different points in the pipeline.

\begin{table*}[t]
\centering
\scriptsize
\setlength{\tabcolsep}{4pt}
\renewcommand{\arraystretch}{1.13}
\caption{Operational proxies for assigning a dominant functional role. The table is not intended as a strict checklist or deterministic decision rule; rather, it provides practical cues for deciding whether a post-training stage is primarily expansion-dominant, reshaping-dominant, or consolidation-dominant in a given setting. Support attrition is listed separately as a diagnostic risk rather than as a fourth dominant role.}
\label{tab:operational_role_proxies}

\begin{tabularx}{\textwidth}{
>{\raggedright\arraybackslash}p{1.65cm}
>{\raggedright\arraybackslash}X
>{\raggedright\arraybackslash}p{2.55cm}
>{\raggedright\arraybackslash}p{3.2cm}
>{\raggedright\arraybackslash}p{2.55cm}
}
\toprule
\textbf{Dominant role}
& \textbf{Typical positive evidence}
& \textbf{Main question answered}
& \textbf{Typical evidence against dominance}
& \textbf{Attrition risk to monitor} \\
\midrule

Expansion
&
Newly occupied effective regions; improved reachability of useful prefixes or locally important actions; emergence of solution patterns that were previously absent or difficult to realize reliably; supervision that exposes the learner to behavior it rarely realized before
&
Does training make previously negligible or inaccessible useful behavior practically reachable?
&
Gains are concentrated mainly on already reachable responses, styles, branches, or state-action choices, with little evidence that new effective regions became available
&
Previously effective regions may shrink while new ones expand; apparent gains may mask losses of earlier capabilities elsewhere
\\

Reshaping
&
Better ranking among already reachable candidates; improved local branch choice on learner-visited states; stronger calibration, alignment, or response quality without clear widening of effective support; probability redistribution within shared effective support
&
Does training improve behavior mainly by reweighting state-action choices that were already effectively available?
&
Improvements depend primarily on importing new trajectories, new prefixes, or previously unseen state-action regions, rather than on improving choice within retained effective support
&
Harmful reweighting inside retained regions may degrade previously reliable behavior even when support is preserved
\\

Consolidation
&
Preservation of gains across stage transitions; successful transfer to a smaller or cheaper student; retention after handoff; amortization of costly search, reflection, or teacher guidance into a reusable policy; robust deployment without the original scaffold
&
Can useful behavior be preserved, transferred, compressed, or amortized across stages, models, or deployment settings?
&
The main gains exist only during the inducing stage and disappear once teacher support, search, or runtime scaffolding is removed
&
Stage transitions may cause support attrition if useful induced behavior is not faithfully retained in the next model or regime
\\

\bottomrule
\end{tabularx}
\end{table*}

\subsection{Hybrid Pipelines as Coordinated Multi-Role Systems}
\label{subsec:framework_hybrid_logic}

Once dominant roles are treated as stage-specific tendencies rather than fixed family labels, hybrid pipelines become structurally natural. Different bottlenecks arise at different points in post-training: some useful behaviors are not yet practically reachable, some failures appear mainly on learner-generated states, and some induced behaviors are costly, fragile, or scaffold-dependent and must later be preserved in a cheaper or more stable form. This is why strong post-training systems increasingly take the form of coordinated multi-stage pipelines rather than isolated single-method recipes \citep{ouyang2022rlhf, rafailov2023dpo, team2025kimi, glm5team2026glm5, liu2025deepseek}.

These bottlenecks are often partially ordered. Useful behavior may first need to be brought into practical reach, then improved under learner-generated rollouts, and finally preserved, transferred, or amortized across handoff. A central systems risk is stage-transition attrition: even when one stage successfully expands support or reshapes behavior, later stages may inadvertently eliminate part of that gain. Section~\ref{sec:hybrid_comparative} examines these issues more directly through stage composition, comparison, and handoff risk.

\subsection{What This Framework Does and Does Not Claim}
\label{subsec:framework_scope_limits}

The framework proposed in this section is a comparative analytic lens rather than a canonical measurement scheme or a complete mechanistic theory. Its notions---including effective support itself---are operational and are intended to organize comparison across major post-training paradigms rather than to supply a uniquely privileged formal vocabulary. Dominant-role assignments should therefore be read as context-dependent analytical summaries, not as immutable family labels, and they do not replace detailed algorithmic, optimization, or systems analysis. Their value is to provide a common language for comparing methods, diagnosing bottlenecks, and reasoning about how stages interact in larger systems. With this framework in place, the next two sections examine off-policy and on-policy post-training as the survey's two top-level intervention regimes, before Section~\ref{sec:hybrid_comparative} returns to their interaction in multi-stage systems.

%% file: sections/offpolicy_posttraining.tex
\section{Off-Policy Post-Training}
\label{sec:offpolicy}

Off-policy post-training improves the model on \emph{externally supplied trajectories}: demonstrations, curated traces, preference data, synthetic teacher outputs, verified solutions, or replayed traces that are not freshly generated by the current learner at the moment of optimization. Under the framework introduced earlier, the provenance axis is therefore fixed: learning remains anchored to externally prepared behavior rather than to learner-generated rollouts. What varies within this regime is the supervision interface and the dominant functional role that different off-policy subfamilies tend to play. Throughout this section, supervision interface is therefore used as a secondary descriptive axis for distinguishing method families that share off-policy provenance but attach learning signals in different ways.

Its main advantage is structural. Because optimization acts on curated behavior rather than on the learner's own sampled states, off-policy learning can expose useful trajectories, prefixes, and local decisions that the learner would not yet reliably realize on its own, while supporting efficient and relatively stable optimization on selected data. Its main limitation is equally structural: off-policy learning does not directly optimize the model on the states it later visits under its own rollout dynamics \citep{bengio2015scheduled, ross2011dagger}. Failures that arise only on learner-generated continuations may therefore remain under-corrected.

\subsection{Supervised Fine-Tuning as Off-Policy Learning}
\label{subsec:offpolicy_sft}

Supervised fine-tuning (SFT) is the most direct and one of the historically dominant forms of off-policy post-training. Given externally supplied instruction-response pairs or other target demonstrations, the model is optimized under teacher forcing to increase the likelihood of target responses. Structurally, SFT combines off-policy provenance with token-target supervision. In modern LLM development, SFT underlies instruction tuning, capability and domain adaptation, safety tuning, and many forms of synthetic-data bootstrapping \citep{wei2022finetuned, sanh2022t0, wang2022supernatural, wang2023selfinstruct, taori2023alpaca, zhou2023lima, li2025demonstration, ye2025limo, singhal2023clinical, chung2024scaling}.

\paragraph{Why SFT belongs to the off-policy regime}
SFT is off-policy in the most direct sense: the learner is trained on externally chosen trajectories rather than on rollouts produced by its current policy. Even when those trajectories are synthetic or produced by an earlier model, they remain externally supplied relative to the learner being updated.

\paragraph{Typical functional role}
SFT does not have a single fixed function. Under the unified framework of this survey, when demonstrations expose trajectory structure or local decisions that the model would not otherwise reliably produce, SFT is often best interpreted as support-expansion-dominant. Representative cases include settings where supervised traces introduce solution structures or reasoning patterns that the learner would not otherwise reliably realize \citep{yu2024metamath, yue2024mammoth, tang2024mathscale, ye2025limo}. When the model already has substantial latent competence, the same supervision interface may instead calibrate, format, align, or select among behaviors that are already partly reachable. In such cases, it is more naturally interpreted here as policy reshaping \citep{ouyang2022rlhf, wang2023selfinstruct, zhou2023lima, dong2023steerlm}. Which role dominates depends not on the token-target interface alone, but on how the target behavior covered by supervision data relates to the learner's existing effective support.

\paragraph{Coverage and failure modes}
What matters in practice is what the supervision data actually add relative to the learner's current effective support. High-quality demonstrations may expose useful prefixes, decompositions, or local action patterns that expand support, whereas demonstrations that mainly reinforce style, format, or preference among already reachable behaviors may primarily reshape response tendencies or local ranking. At the same time, SFT is especially vulnerable when teacher-forced training traces differ sharply from the learner's eventual rollout states. In such cases, behavior that appears robust under curated prefixes may prove brittle under self-generated continuations \citep{bengio2015scheduled, lamb2016professor}. SFT can also overfit narrow trajectory styles, import brittle rationales, or amplify superficial regularities when supervision is too narrow or poorly aligned with deployment.

\subsection{Offline Preference Optimization}
\label{subsec:offpolicy_preference}

Preference optimization methods improve the model using externally supplied preference comparisons, ranked outputs, or preference-derived surrogates. In their standard formulations, these signals are collected or constructed offline, and the learner is updated against a fixed comparison dataset rather than on fresh rollouts sampled from the current policy during each optimization step. Their characteristic supervision interface is therefore pairwise or ranked preference supervision attached to offline candidate comparisons. The discussion here concerns that standard offline setting.

\paragraph{Typical functional role}
Preference optimization often acts primarily by reweighting behavior within regions that are already effectively available. Chosen and rejected responses are usually drawn from behavior the model can already produce, and the training signal encourages the learner to increase the relative probability of preferred continuations while suppressing less desirable ones. In this sense, preference optimization is typically best interpreted as off-policy reshaping: it improves ranking, style, calibration, alignment, or response quality within behavior that is already effectively available \citep{rafailov2023dpo, ethayarajh2024kto, hong2024orpo, meng2024simpo, tang2024gpo}.

Preferred examples can still contribute to support expansion when they expose trajectory structures the learner would not otherwise reliably realize, especially when they come from substantially stronger generators. Even so, in most standard offline formulations the dominant role remains reshaping rather than broad expansion.

\paragraph{Why preference optimization is attractive in practice}
Preference optimization occupies an important practical middle ground. Compared with plain SFT, it provides a more explicit mechanism for enforcing ranking and alignment among multiple candidate behaviors. Compared with on-policy RL, standard offline preference optimization is often easier to train and more computationally efficient in practice \citep{rafailov2023dpo}, because it avoids repeated online rollout and reward estimation during each update, which are central components of RLHF-style training pipelines \citep{ouyang2022rlhf}. Its main weakness is structural: supervision is limited to the comparison space that offline data collection makes visible. If important failures arise only on learner-generated prefixes, preference optimization may improve relative ranking among curated responses while leaving rollout-dependent errors largely untouched.

Online preference optimization forms an important bridge to the on-policy regime; we return to it in Section~\ref{subsec:onpolicy_online_preference}.

\subsection{Off-Policy Teacher-Guided Transfer}
\label{subsec:offpolicy_teacher_transfer}

Off-policy teacher-guided transfer refers to settings in which the student is trained on teacher-provided or teacher-mediated traces that remain externally supplied relative to the current learner. Prominent examples include classic distillation on teacher outputs, as well as reasoning-oriented variants such as rationale distillation, explanation transfer, and other forms of offline teacher-guided reasoning transfer \citep{hinton2015distilling, mukherjee2023orca, mitra2024orca2, hsieh2023distilling, shridhar2023distilling, snell2022context, huang2022icl_distill}. Under the present framework, these methods do not define a new provenance regime; they are best understood as a teacher-guided supervision interface applied to externally supplied traces.

Off-policy distillation can in principle contribute to either expansion or reshaping, depending on what the teacher exposes. In modern systems, however, its most characteristic systems-level role is behavioral consolidation. The point is often not merely to produce better behavior in the moment, but to preserve, transfer, compress, or amortize behavior that was first induced elsewhere—for example by a larger teacher, an RL stage, a search process, or an expensive reflective pipeline. The main risk is incomplete transfer: a student may match teacher outputs on the distillation distribution without fully internalizing the behavior that made those outputs reliable.

\subsection{A Compact Survey of Off-Policy Subfamilies}
\label{subsec:offpolicy_compact_table}

Table~\ref{tab:offpolicy_family_summary} plays a comparative role for the off-policy regime, aligning supervision interface, usual dominant role, best-fit use case, and characteristic limitation across the main off-policy subfamilies.

\begin{table*}[t]
\raggedright
\scriptsize
\setlength{\tabcolsep}{4pt}
\renewcommand{\arraystretch}{1.15}
\caption{Compact survey of major off-policy subfamilies under the unified lens. The table highlights their typical supervision interfaces, usual dominant role, and most characteristic limitation.}
\label{tab:offpolicy_family_summary}
\begin{adjustbox}{max width=\textwidth}
\begin{tabular}{>{\raggedright\arraybackslash}p{2.1cm}
                >{\raggedright\arraybackslash}p{3.0cm}
                >{\raggedright\arraybackslash}p{3.0cm}
                >{\raggedright\arraybackslash}p{3.45cm}
                >{\raggedright\arraybackslash}p{3.7cm}}
\toprule
\textbf{Subfamily} & \textbf{Supervision interface} & \textbf{Usual dominant role} & \textbf{Best suited for} & \textbf{Common limitation} \\
\midrule

SFT
& Token targets on demonstrations, curated responses, or synthetic reasoning traces
& Support expansion or reshaping, depending on supervision coverage relative to existing effective support
& Importing trajectory structure, exposing latent capability, formatting, alignment, and domain adaptation
& Mismatch between teacher-forced prefixes and learner rollout states; weak direct correction on learner-specific failures \\

Offline preference optimization
& Preference pairs, rankings, or chosen/rejected comparisons
& Usually reshaping
& Offline alignment, ranking, response quality control, and calibration among reachable behaviors
& Curated candidate space may miss rollout-dependent failures; limited direct correction on learner-generated states \\

Off-policy teacher-guided transfer
& Teacher-guided targets, logits, filtered traces, or rationale transfer
& Often behavioral consolidation
& Transfer, retention, compression, amortization, and deployment-oriented handoff
& Incomplete transfer when teacher behavior depends on privileged context or costly scaffolding \\

\bottomrule
\end{tabular}
\end{adjustbox}
\end{table*}

\subsection{Section Summary: Where Off-Policy Learning Helps Most—and Where It Stops}
\label{subsec:offpolicy_takeaways}

Across its subfamilies, off-policy learning is strongest when useful behavior must be externally introduced, broadly established, or efficiently reshaped on curated data. This makes it especially valuable for early behavior establishment, broad off-policy support, offline quality control, and later consolidation into more stable or deployable forms.

Its structural blind spot is learner-state mismatch. Because optimization is driven by externally supplied trajectories rather than by the learner's own rollout states, purely offline supervision may improve the training distribution while leaving deployment-relevant failures under-corrected. This is especially true when errors depend on self-generated branching, compounding intermediate mistakes, or recovery from prefixes that curated data rarely visits. The issue is therefore regime-level rather than specific to any single off-policy subfamily. Its practical effectiveness also depends heavily on the breadth, quality, and deployment relevance of the external data source: when curated traces are too narrow, weakly matched, or superficially filtered, off-policy updates may reshape visible behavior without producing robust gains under actual rollout conditions. More precisely, familiar criticisms of narrow off-policy tuning are often better interpreted as consequences of deployment mismatch and overly narrow reshaping, rather than as evidence that externally supplied supervision is intrinsically incapable of improving behavior.

For that reason, off-policy stages often serve as establishment, shaping, or consolidation stages within a larger pipeline, while later on-policy stages are frequently needed for learner-state correction once behavior has become at least partly reachable under the learner's own policy. In practice, strong behavior is often more robust when early off-policy import is followed by either on-policy correction or especially careful handoff design across stages.

%% file: sections/onpolicy_posttraining.tex
\section{On-Policy Post-Training}
\label{sec:onpolicy}

On-policy post-training improves the model on learner-generated rollouts: trajectories freshly sampled from the current policy and then evaluated, revised, filtered, or otherwise used to construct the next update. Under the framework of this survey, the provenance axis is therefore fixed: learning is anchored to the learner's own visited states rather than to externally prepared traces. What varies within this regime is the supervision interface and the dominant functional role through which different on-policy subfamilies improve behavior. Accordingly, the discussion below uses supervision interface as a secondary descriptive axis for distinguishing methods that share learner-generated rollouts but differ in how feedback is attached to those visited states.

Its main advantage is structural. Because optimization is driven by trajectories the learner actually generates, on-policy learning can observe failures that arise only after self-generated branching, compounding rollout error, or interaction-specific state transitions. It is therefore especially valuable when the main bottleneck is not merely lack of external supervision, but the need to correct behavior on the states the learner itself visits at deployment time. Its main limitation is equally structural: on-policy optimization does not by itself reliably import useful behavior that lies outside the learner's current effective support.

\subsection{RLHF, RLAIF, and RLVR as On-Policy Learning}
\label{subsec:onpolicy_rlhf_rlvr}

The most familiar form of on-policy post-training is reinforcement learning on fresh model rollouts. In RLHF, a reward model or related evaluator scores responses sampled from the current policy, and the policy is updated to increase expected return \citep{ouyang2022rlhf, bai2022hh_rlhf, stiennon2020summarize}. In constitutional or AI-feedback variants, often grouped under reinforcement learning from AI feedback (RLAIF), the evaluative signal may come from rule-based critique or AI-generated preference and revision mechanisms rather than purely human labels \citep{bai2022constitutional, lee2023rlaif_vs_rlhf}. In RLVR and related reasoning-focused systems, the signal is instead derived from verifiable outcomes, tests, or structured correctness checks applied to learner-generated traces \citep{shao2024deepseekmath, deepseek2025r1, yu2025dapo, wen2026RLVR, zheng2025gspo}.

\paragraph{Why RLHF/RLVR belongs to the on-policy regime}
These methods are on-policy because updates are computed on fresh trajectories sampled from the current learner rather than on externally curated candidate sets. This is the structural distinction from offline preference optimization or reward-ranked supervision, even when the supervision interface may look similar.

\paragraph{Typical functional role}
Under the unified framework of this survey, on-policy RL is often best interpreted as reshaping-dominant. It typically improves behavior on states the learner already visits by changing local branch choice, strengthening recovery from intermediate mistakes, and increasing the probability of verifier-consistent or otherwise higher-return continuations within rollout regions that are already effectively available.

Selective support expansion can still occur, but the clearest cases usually arise under structured rollout guidance rather than under unconstrained rollout optimization alone. If a useful branch is already within the learner's effective support, then making it more reliable is better understood as reshaping. By contrast, selective expansion can occur when guidance mechanisms---such as self-hints, scaffolded prompting, structured templates, or other strong intermediate guidance---alter the rollout distribution so that the learner repeatedly encounters behavior it would not otherwise reliably realize under its unassisted policy \citep{liao2026sage, expo2025, wu2025templaterl, hipo2026, zhou2025ScaffoldedRL}.

\paragraph{Representative failure modes}
The main risks are evaluator sensitivity and optimization instability. Reward models, verifier signals, or constitutional rules may imperfectly reflect true task quality and can therefore induce reward hacking, brittle overoptimization, or calibration drift when the learner discovers shortcuts that score well without genuinely improving behavior \citep{casper2023open, skalse2022defining, gao2023scaling}. On-policy RL is also expensive and often harder to stabilize than offline supervision because each training cycle depends on fresh learner rollouts, repeated evaluation of those rollouts, and policy updates on the resulting on-policy trajectories \citep{zheng2023pposecrets, zheng2025gspo}. Large or poorly calibrated updates can also compromise previously useful behavior outside the optimized rollout region, leading to support attrition or retention failures rather than uniform improvement \citep{sabbaghi2026robust}.

\subsection{Online Preference Optimization}
\label{subsec:onpolicy_online_preference}

Online preference optimization retains the preference-based supervision interface of offline preference optimization, but moves preference updates closer to data refreshed from the current policy \citep{Calandriello2024onlinePO, qi2024ofsdpo, shi2025onlinedpo, bai2025copo}. In the present framework, it therefore occupies an intermediate position: the supervision signal remains preference-based, while the trajectory provenance shifts toward learner-generated or current-policy data.

Its role is still more naturally understood as reshaping rather than broad support expansion. Compared with standard offline preference optimization, online variants can better expose ranking failures that are less visible when preference updates rely on fixed offline comparisons, because updates are tied more closely to current-policy data. This makes online preference optimization a useful bridge between offline preference optimization and more explicitly on-policy reinforcement learning.

Because it moves preference updates closer to learner-generated data, online preference optimization remains constrained by what the learner actually produces. It can improve alignment among learner-relevant candidates, but it still depends on what the learner actually generates and therefore cannot by itself overcome broader exploration limitations when desirable behaviors remain rarely visited \citep{bai2025copo}. Its practical strength therefore lies in bringing preference-based supervision closer to learner-generated states without taking the full form of reward-driven reinforcement learning.

\subsection{On-Policy Process Supervision and Verifier-Guided Correction}
\label{subsec:onpolicy_process_verifier}

Another major on-policy region consists of methods that attach feedback to intermediate learner-generated states rather than only to final trajectories. This family includes process supervision, step-level verifier guidance, intermediate critique signals, and related forms of rollout-time correction in which the learner's own evolving prefixes are evaluated and improved during or after generation \citep{wang2024mathshepherd, setlur2025rewarding, zhang2025rewardSql, dai2024lineLevelPrmCode}. The supervision interface here differs from scalar terminal reward: feedback is attached more locally to the learner's trajectory, often at intermediate reasoning states or branch points.

Under the unified framework of this survey, these methods are usually best understood as fine-grained learner-state reshaping. Their main contribution is to improve local branch choice on learner-visited prefixes, enable earlier correction of intermediate mistakes, and provide denser credit assignment than end-of-trajectory reward alone. Process-level guidance is especially useful in long-horizon reasoning, code synthesis, and tool-use or agentic interaction settings, where local errors can propagate and accumulate into final failure \citep{setlur2025rewarding, wang2024mathshepherd, dai2024lineLevelPrmCode, xi2025agentprm}.

Its main risks are verifier or process mismatch, the scalability of process supervision, and reachable-state dependence, since only states and branches the learner actually visits can be corrected. If the intermediate evaluator is weak, shallow, or poorly aligned with final task quality, dense correction may simply amplify the wrong local signals.

\subsection{Iterative Refinement, Self-Improvement, and Other On-Policy-Like Improvement Loops}
\label{subsec:onpolicy_iterative}

A broader class of on-policy or strongly on-policy-like methods improves the model through iterative refinement loops built around learner-generated behavior. This family includes critique--revise loops, reasoning self-training procedures, self-correction pipelines, and related methods that repeatedly generate learner behavior, assess it, revise or filter it, and then feed the resulting traces back into subsequent updates or refinement rounds \citep{singh2023restem, zhang2025critiquegrpo, ma2025s2r, kumar2025self-correct, pang2024self-improvement, bao2025fixing, zelikman2022star, gulcehre2023rest, zhang2024rest-mcts}. What unifies these methods is not a single supervision interface, but the fact that improvement remains centered on fresh learner behavior and the iterative loop constructed around it, even when some traces are later cached, filtered, or reused.

In the common case, these methods are best viewed as reshaping-dominant: the learner already produces candidate trajectories, and the loop improves selection, correction, decomposition, or revision on those branches. A recurring risk is that weak critique or revision signals may simply reinforce superficial patterns rather than produce genuinely better behavior. Another is that gains achieved through repeated refinement may depend on the loop itself and therefore fail to transfer cleanly into the final deployment policy.

\subsection{On-Policy Teacher-Guided Transfer}
\label{subsec:onpolicy_teacher_transfer}

On-policy teacher-guided transfer attaches teacher-derived targets or corrections directly to learner-generated rollouts. The learner first produces its own behavior, and a teacher or privileged evaluator then supplies improved actions, targets, critiques, or corrected continuations on those visited states \citep{gu2024minillm, agarwal2024opd, ye2025blackbox_opd, ye2026opcd, zhao2026selfDistilledReasoner, yang2026gopd, sang2026crisp}. The defining feature is therefore not a separate provenance regime, but teacher-guided supervision applied on the learner's own rollout distribution.

In its most characteristic form, on-policy teacher-guided transfer is best interpreted as reshaping-dominant learner-state correction. Because the teacher acts on states the student actually visits, the main effect is often to improve local decisions, branch selection, or recovery within learner-reached regions rather than to import wholly external trajectories. At the systems level, this family also connects naturally to behavioral consolidation: when online gains are later distilled or amortized into a more stable student, teacher guidance can help preserve behavior discovered on learner-generated rollouts.

Teacher-guided correction can also support selective expansion when corrected local actions steer rollouts into downstream prefixes and state-action patterns that were previously outside the learner's effective support under its unaided policy. The main practical risks are teacher cost, limited teacher advantage, and incomplete retention after teacher removal.

\subsection{A Compact Survey of On-Policy Subfamilies}
\label{subsec:onpolicy_compact}

Table~\ref{tab:onpolicy_family_summary} plays the same comparative role for the on-policy regime, aligning supervision interface, usual dominant role, best-fit use case, and characteristic limitation across the main on-policy subfamilies, with particular emphasis on how they differ in learner-state correction.

\begin{table*}[t]
\raggedright
\scriptsize
\setlength{\tabcolsep}{3.8pt}
\renewcommand{\arraystretch}{1.14}
\caption{Compact survey of major on-policy subfamilies under the unified lens. The table emphasizes typical supervision interfaces, usual dominant role, and the most characteristic limitation.}
\label{tab:onpolicy_family_summary}
\begin{adjustbox}{max width=\textwidth}
\begin{tabular}{>{\raggedright\arraybackslash}p{2.0cm}
                >{\raggedright\arraybackslash}p{2.85cm}
                >{\raggedright\arraybackslash}p{2.85cm}
                >{\raggedright\arraybackslash}p{3.4cm}
                >{\raggedright\arraybackslash}p{3.7cm}}
\toprule
\textbf{Subfamily} & \textbf{Supervision interface} & \textbf{Usual dominant role} & \textbf{Best suited for} & \textbf{Common limitation} \\
\midrule

RLHF / RLAIF / RLVR
& Rewards, evaluators, tests, or constitutional / AI-feedback on fresh learner rollouts
& Usually reshaping; sometimes selective expansion under strong rollout guidance
& Direct correction of rollout-time failures, preference-sensitive behavior, and verifiable reasoning outcomes
& High cost, evaluator mismatch, optimization instability, and limited import of behavior not yet reachable \\

Online preference optimization
& Preferences refreshed from current-policy candidates or periodically updated comparison data
& Usually reshaping
& Bringing preference-based supervision closer to learner-generated states
& Limited by what the learner generates, especially under sparse refresh or rare desirable behaviors \\

Process / verifier-guided correction
& Step-level process feedback, intermediate verifiers, or local critique on learner trajectories
& Fine-grained learner-state reshaping
& Denser local correction in reasoning, coding, and long-horizon tasks with intermediate failure points
& Verifier/process mismatch, scalability, reachable-state dependence, and proxy overoptimization \\

Iterative refinement / self-improvement
& Critique, revision, reranking, or repeated refinement loops on fresh learner behavior
& Usually reshaping
& Improving decomposition, revision quality, branch choice, and self-correction on learner-generated candidates
& Weak critique may reinforce superficial patterns, and gains from repeated refinement may transfer poorly to the final deployment policy \\

On-policy teacher-guided transfer
& Teacher targets, corrections, or privileged feedback attached to learner rollouts
& Usually reshaping; systems-level link to behavioral consolidation
& Student-state-aligned transfer, online correction, and amortization of rollout-time gains
& Teacher cost, limited teacher advantage, and incomplete retention after teacher removal \\

\bottomrule
\end{tabular}
\end{adjustbox}
\end{table*}

\subsection{Section Summary: Learner-State Correction Without Automatic Behavior Import}
\label{subsec:onpolicy_takeaways}

Across its subfamilies, on-policy learning is strongest when the main bottleneck is failure on learner-generated states. Because optimization occurs on fresh learner rollouts, on-policy methods can directly observe compounding reasoning errors, bad branch selection, weak recovery from intermediate mistakes, and other rollout-dependent failures that curated offline data may only weakly expose.

Its main limitation is complementary: on-policy learning does not by itself reliably import useful behavior that the learner almost never visits. If desirable solution branches, decompositions, or local action patterns rarely appear in fresh learner rollouts, on-policy correction has little to reinforce unless exploration, scaffolding, or earlier off-policy stages first bring those behaviors into view.

This is why on-policy stages often refine and stabilize behavior that earlier off-policy stages helped make reachable, even though guided variants can sometimes support selective expansion. In practice, this also means that on-policy learning remains sensitive to exploration quality, evaluator reliability, and rollout cost: when useful behaviors do not appear in learner rollouts, or when evaluative signals are noisy or unstable, on-policy correction alone may have little solid ground on which to build. Accordingly, improved performance on learner-generated states does not by itself imply broad import of behaviors that were absent from the learner's rollout distribution to begin with. Under the unified framework, on-policy learning is therefore best understood as a complementary regime whose value depends on whether useful behavior is already at least partly reachable under the learner's own policy.

%% file: sections/hybrid_comparative_analysis.tex
\section{Hybrid Pipelines and Comparative Analysis}
\label{sec:hybrid_comparative}

The previous sections reviewed off-policy and on-policy learning as the survey's two top-level regimes. Strong LLM systems, however, are increasingly built by coordinating multiple stages across a pipeline rather than by choosing one regime in isolation. This section therefore treats hybrid pipelines as systems-level compositions of different provenances, interfaces, and dominant roles, and then uses that view to compare major paradigms more directly. Here, supervision interfaces remain a secondary comparative layer rather than a separate organizing axis: they help explain why stages with similar provenance can still differ materially in learning dynamics, handoff behavior, and consolidation difficulty.

\subsection{Why Hybrid Pipelines Are Structurally Natural}
\label{subsec:hybrid_comparative_why_needed}

Single-regime thinking is often insufficient because different post-training bottlenecks arise at different stages of behavior acquisition, correction, and stabilization. Off-policy stages are often better suited to exposing or importing useful behavior that the learner would not yet reliably realize on its own, on-policy stages are often better suited to correcting failures on learner-generated states, and later stages may still need to preserve, transfer, amortize, or compress induced behavior across handoff. No single regime addresses all of these bottlenecks equally well. Hybrid pipelines therefore arise not merely because practitioners stack methods opportunistically, but because strong systems often need to coordinate support expansion, learner-state reshaping, and behavioral consolidation across stage boundaries.

\subsection{Stage Composition: Expansion, Reshaping, and Consolidation}
\label{subsec:hybrid_comparative_three_stage_logic}

Although real systems vary substantially, many strong pipelines can still be interpreted through a recurring three-stage logic:
\begin{enumerate}[leftmargin=2em]
    \item \textbf{Support expansion:} off-policy supervision exposes useful traces, decompositions, intermediate reasoning structures, or solution patterns that the learner would not yet reliably generate on its own;
    \item \textbf{Policy reshaping on learner-generated states:} on-policy optimization improves behavior on the states the model actually visits, correcting rollout-specific failures that external data alone may not expose;
    \item \textbf{Behavioral consolidation:} useful behavior induced by earlier stages is stabilized, preserved, transferred, or amortized into a model form that can be reused or deployed efficiently.
\end{enumerate}

This logic is functional rather than procedural: real systems may collapse, interleave, or iterate stages, and the first stage may range from strong support expansion to weaker forms of off-policy behavior establishment or reshaping. The main comparative question is whether stage order is aligned with the current bottleneck: insufficient effective support, poor learner-state behavior, or weak behavioral preservation across handoff.

Table~\ref{tab:hybrid_stage_patterns_summary} summarizes recurring stage-composition patterns and highlights their typical bottlenecks and handoff risks; several especially common patterns are discussed below.

\begin{table*}[t]
\centering
\scriptsize
\setlength{\tabcolsep}{4.0pt}
\renewcommand{\arraystretch}{1.13}
\caption{Typical stage-composition patterns in modern post-training pipelines, including both cross-regime pipelines and important within-regime multi-stage compositions. The final two columns summarize the main bottleneck each pattern addresses and the characteristic handoff failure that arises when stage transitions are miscalibrated.}
\label{tab:hybrid_stage_patterns_summary}
\begin{adjustbox}{max width=\textwidth}
\begin{tabular}{>{\raggedright\arraybackslash}p{2.3cm} >{\raggedright\arraybackslash}p{2.5cm} >{\raggedright\arraybackslash}p{2.4cm} >{\raggedright\arraybackslash}p{2.3cm} >{\raggedright\arraybackslash}p{2.7cm} >{\raggedright\arraybackslash}p{2.8cm}}
\toprule
\textbf{Pipeline pattern} & \textbf{Early support stage} & \textbf{Reshaping stage} & \textbf{Consolidation stage} & \textbf{Main bottleneck addressed} & \textbf{Common failure if miscalibrated} \\
\midrule
SFT $\rightarrow$ preference optimization
& Supervised exposure to useful traces, decompositions, or solution patterns
& Preference-based reweighting within already reachable behavior
& Often implicit or omitted
& Broad behavior establishment first; preference-level calibration later
& Narrow style or ranking over-optimization while rollout-specific failures remain unresolved \\

SFT $\rightarrow$ RLHF / RLVR
& External demonstrations or synthetic traces provide initial support for useful behavior
& On-policy correction improves what the learner actually does under its own rollout dynamics
& Optional later transfer or policy refinement
& Initial capability exposure followed by learner-state correction
& Weak early support raises RL cost and instability; overfit early support limits later RL generalization \\

Interleaved SFT--RL training
& Off-policy supervision continuously maintains behavior the learner would not yet discover alone
& On-policy rollouts simultaneously correct learner-state failures
& Often implicit or deferred
& Continuous support plus learner-state correction without brittle stage handoff
& Poor coupling or gradient interference destabilizes joint supervision--RL training \\

SFT $\rightarrow$ RL $\rightarrow$ distillation
& External supervision provides early support for useful behavior
& On-policy RL or search improves rollout-time behavior
& Distillation transfers or amortizes the resulting behavior into a cheaper or more stable student
& Consolidate costly or specialist-induced behavior into reusable competence
& Specialist gains are not preserved after transfer or merging \\

Search / verifier-heavy induction $\rightarrow$ policy transfer
& Search, filtering, or verifier-heavy procedures induce strong but costly behavior
& Optional learner-state correction or policy improvement
& Distillation or policy transfer compresses behavior into a reusable policy
& Convert search- or verifier-induced behavior into deployable capability
& Student imitates outputs superficially but loses robustness once search is removed \\
\bottomrule
\end{tabular}
\end{adjustbox}
\end{table*}

\paragraph{SFT $\rightarrow$ preference optimization}
A common pattern is to begin with supervised fine-tuning or instruction tuning to establish broad useful behavior, then apply preference optimization to reshape ranking, style, safety, or output quality within an already reachable region \citep{rafailov2023dpo, meng2024simpo, Calandriello2024onlinePO, grattafiori2024llama, lambert2025tulu}. Depending on the setting, the SFT stage may contribute either to support expansion or to policy reshaping that makes the learner's initial responses more usable. The later preference-optimization stage is then not redundant with SFT: it refines relative ranking and calibration among behaviors that the earlier stage has already made sufficiently available or usable.

\paragraph{SFT $\rightarrow$ RLHF / RLVR}
Another common pattern is to use SFT as an initial support-providing stage and then apply on-policy RL to improve learner-time performance under preference, constitutional, or verifiable reward signals \citep{ouyang2022rlhf, bai2022constitutional, shao2024deepseekmath, deepseek2025r1}. Depending on the setting, the SFT stage may contribute either to support expansion or to policy reshaping that makes the learner's initial rollouts more usable. The later RL stage is then not redundant with the SFT stage: it addresses learner-state failures that become visible only once the learner acts under its own policy.

\paragraph{Interleaved or joint SFT--RL training}
Some systems interleave or unify off-policy supervision with on-policy reshaping rather than separating them into two disjoint stages \citep{ma2026interleave, yan2025luffy, fu2025srft, liu2025uft, zhang2025chord, wang2025grao}. Their shared motivation is that a hard SFT $\rightarrow$ RL handoff can be brittle: off-policy supervision helps provide traces, decompositions, or other useful behaviors that purely on-policy optimization may not reliably discover, whereas RL is better at refining behavior on learner-generated states. Under the unified framework, these methods are best understood as attempts to soften the transition between early off-policy support and learner-state reshaping.

\paragraph{SFT $\rightarrow$ RL $\rightarrow$ distillation}
This pattern is increasingly important in reasoning- and agent-heavy systems. External supervision first provides early support for useful behavior, domain- or task-specific RL then strengthens behavior under rollout, and a later distillation or cross-stage transfer stage consolidates these specialist capabilities into a cheaper, more stable, or more general deployment model \citep{deepseek2025r1, glm5team2026glm5, xiao2026mimo, yang2026nemotron}. In many recent systems, different experts are trained for different domains or interaction modes and then distilled or merged into a final unified model. Under the unified framework, the key challenge is not only discovering useful behavior, but preserving and consolidating it across domains and stage boundaries.

\paragraph{Search- or verifier-heavy induction $\rightarrow$ policy transfer}
A recurring systems pattern is to first induce strong but costly behavior through search- or verifier-heavy procedures, and then transfer or distill that behavior into a more direct deployment policy \citep{luo2026agentark, dixit2026aletheia, li2025mixtureteacher}. Related work on process supervision and verifier-guided reasoning makes clear why such transfer can be necessary in the first place: the inducing machinery may be powerful, but also too expensive, fragile, or annotation-intensive to serve directly as the final deployed policy \citep{lightman2023lets, wang2024mathshepherd}. Under the unified framework, these pipelines separate behavior induction from later consolidation, shifting expensive test-time or verifier-mediated competence into a cheaper reusable model form.

\subsection{Why Consolidation Becomes Central in Hybrid Pipelines}
\label{subsec:hybrid_comparative_consolidation}

In hybrid pipelines, useful behavior often first appears in a costly, fragile, scaffold-dependent, or domain-specialized form. Post-training is therefore not finished when such behavior is first induced, but only when it can survive later stages and remain available under the intended deployment regime. Consolidation becomes central not because it defines a separate learning regime, but because it addresses a distinct systems bottleneck: preserving, transferring, amortizing, or compressing useful behavior across stage boundaries.

Distillation is especially prominent here because it is a common mechanism for carrying induced behavior into a cheaper, more stable, or more deployable student \citep{agarwal2024opd, ye2026opcd, zhao2026selfDistilledReasoner, luo2026agentark, glm5team2026glm5}. Under this view, consolidation is not an optional final polishing step, but the systems-level response to the risk that induced behavior will otherwise be lost, diluted, or left dependent on costly runtime scaffolds.

\subsection{Attrition-Aware Failure Modes in Hybrid Pipelines}
\label{subsec:hybrid_comparative_failure_modes}

Hybrid pipelines are powerful, but they also introduce failure modes that are best understood as problems of stage interaction and behavior retention. A pipeline may fail not because an earlier stage never induced useful behavior, but because later stages fail to preserve it, reshape it under incompatible assumptions, or induce support attrition during handoff.

\paragraph{Stage mismatch}
A behavior induced in one stage may depend on assumptions that no longer hold in the next. For example, trajectories imported by off-policy supervision may not remain stable once the learner begins rolling out under its own policy, and behavior improved under RL may not survive transfer into a later student or deployment regime. The issue here is not simple undertraining, but mismatch between the conditions under which behavior is induced and those under which it must later operate.

\paragraph{Stage-transition calibration}
A hybrid pipeline may fail not because any single stage is ineffective, but because the transition between stages is poorly calibrated. If an earlier off-policy stage is too weak, useful behavior may not yet be sufficiently reachable, making later on-policy correction inefficient or unstable. If it is too strong, too narrow, or too specialized, later RL may inherit brittle patterns, reduced generalization headroom, or optimization biases that are difficult to undo.

\paragraph{Gradient interference across stages}
When SFT-like and RL-like updates are alternated or jointly optimized, their update dynamics may differ substantially, creating optimization interference across stages. Large supervised updates can overwrite fine-grained improvements discovered by RL, whereas weak RL updates may fail to redirect behavior established by earlier supervision. In such cases, useful behavior is not merely hard to discover, but hard to preserve under mismatched update dynamics \citep{yuan2025mitigating, zhao2026prism}.

\paragraph{Transfer attrition and preservation failure}
Useful behavior may be attenuated, overcompressed, or selectively lost when transferred across stages, especially when a strong but expensive policy is distilled into a cheaper model or when later adaptation aggressively rewrites previously reliable competence. In the language of Section~\ref{subsec:functional_roles}, the issue is not only whether behavior is reshaped, but whether previously effective regions are lost during consolidation and handoff.

\paragraph{Evaluator inconsistency}
Hybrid systems may rely on multiple evaluators across stages---human preference, reward models, verifiers, constitutions, teachers, or filtered search outputs. If these signals are misaligned, stage composition can become brittle even when each stage appears reasonable in isolation, because later stages may reinforce behavior that earlier stages were implicitly trying to suppress.

These failure modes do not argue against hybrid pipelines. Rather, they show that stage composition is itself a scientific problem, not merely an implementation detail. Major post-training paradigms must therefore also be compared in terms of the bottlenecks they address and the handoff risks they are structurally positioned to create or mitigate.

\subsection{Comparative Analysis Through the Unified Lens}
\label{subsec:hybrid_comparative_unified_lens}

We now compare major post-training paradigms through the unified framework developed in Section~\ref{sec:unified_framework}. Building on the preceding discussion of stage composition, consolidation, and attrition, the goal is not to rank methods by a single scalar notion of quality, but to clarify what different methods are structurally positioned to do, what kinds of failures they are most directly positioned to correct, and why they so often complement rather than replace one another.

\subsubsection{Why Trajectory Provenance Is the Most Informative Top-Level Axis}
\label{subsubsec:comparative_provenance}

For the comparative goals of this survey, trajectory provenance remains the most informative top-level axis because it determines where optimization actually occurs. Methods trained on externally supplied trajectories are structurally well positioned to import behaviors that the learner would not yet reliably realize on its own, whereas methods trained on learner-generated rollouts are better positioned to correct failures that arise on the learner's own visited states. This is why objective family alone is often too coarse for comparison: similar interfaces, such as preferences, verifiers, or teacher guidance, can appear in either regime, but their practical role changes once the optimized trajectories differ.

\subsubsection{Distribution-Level Comparison: Expansion and Reshaping}
\label{subsubsec:comparative_expansion_reshaping}

At the distribution level, the key contrast is whether a method is better positioned to bring useful behavior into practical reach or to improve behavior within regions the learner already visits. Under many common post-training settings, off-policy methods are often better positioned to support expansion, whereas on-policy methods are more often reshaping-dominant because they act directly on learner-generated states. This comparison should still be read as a statement about dominant tendencies rather than fixed labels: the same method family may play different roles depending on whether the main bottleneck is missing reachable behavior or poor choice within already reachable behavior. Local action-level improvement can still have downstream trajectory-level consequences, so reshaping and selective support expansion need not be perfectly separable in autoregressive settings.

\subsubsection{Systems-Level Comparison: Why Consolidation Matters}
\label{subsubsec:comparative_consolidation}

A purely distribution-level comparison is still incomplete because major post-training paradigms also differ in how well they preserve useful behavior across stage transitions, model transfer, and deployment handoff. This is where behavioral consolidation becomes comparative rather than merely definitional: some stages mainly induce or refine behavior, while others are especially important for retaining, transferring, amortizing, or compressing that behavior into a more deployable form. In comparative terms, off-policy transfer is often stronger when the main goal is to carry expensive induced behavior into a cheaper student, whereas on-policy teacher-guided transfer is often stronger when preservation must remain aligned with the learner's own rollout states. For multi-stage systems, the central question is therefore not only what a method can improve in isolation, but also how reliably those gains survive later handoff.

\subsubsection{How Major Method Families Compare Under This Lens}

Across major method families, the most informative differences do not lie only in objective names, but in the combination of trajectory provenance, typical train states, supervision interface, and stage-transition risk. Some families are structurally stronger at importing or establishing useful behavior, others at correcting learner-state failures, and others at preserving induced behavior across handoff. Table~\ref{tab:comparative_unified_lens_merged} summarizes these dominant tendencies under the unified lens of trajectory provenance, supervision interface, functional role, and stage-transition risk.

\begin{table*}[t]
\centering
\tiny
\setlength{\tabcolsep}{3.2pt}
\renewcommand{\arraystretch}{1.04}
\caption{Comparison of major post-training paradigms under the unified lens. Entries summarize dominant tendencies rather than exhaustive or mutually exclusive assignments. Functional role refers to typical dominant usage; the boundary-note column indicates when the family-level label is most likely to weaken.}
\label{tab:comparative_unified_lens_merged}

\newcolumntype{L}[1]{>{\raggedright\arraybackslash}p{#1}}
\newcolumntype{Y}{>{\raggedright\arraybackslash}X}

\begin{tabularx}{\textwidth}{L{1.55cm} L{1.0cm} L{1.45cm} L{1.55cm} L{1.85cm} L{2.2cm} L{2.2cm} Y}
\toprule
\textbf{Method family} & \textbf{Prov.} & \textbf{Typical train states} & \textbf{Interface} & \textbf{Dominant role} & \textbf{Boundary note} & \textbf{Main strength} & \textbf{Typical attrition / handoff risk} \\
\midrule

SFT
& Off-pol.
& Curated prompts and prefixes
& Token targets
& Expansion or reshaping
& Weakens under staged synthetic rewriting or strong teacher pipelines
& Efficient behavior import; strong format alignment
& Exposure mismatch and forgetting of previously effective behavior \\

Offline preference optimization
& Usually off-pol.
& Curated chosen/rejected responses
& Ranked preferences
& Usually reshaping
& Weakens under online preference refresh
& Strong offline ranking and alignment calibration
& Weak learner-state correction; may over-optimize style while rollout failures remain \\

Off-policy teacher-guided transfer
& Off-pol.
& Teacher traces or logits
& Teacher-guided targets
& Usually consolidation
& Weakens when teacher guidance is attached to learner rollouts
& Efficient teacher transfer and amortization
& Teacher--deployment mismatch; subtle behavior may be lost in compression \\

On-policy RLHF / RLAIF / RLVR
& On-pol.
& Learner rollouts
& Rewards, evaluators, tests
& Learner-state reshaping
& Weakens when replay or scaffolds dominate training states
& Direct correction of rollout-time failures
& Expensive, evaluator-sensitive, and hard to preserve after handoff \\

Online preference optimization
& Mixed / on-pol.
& Current-policy preference candidates
& Ranked preferences
& Usually reshaping
& Weakens when data stay mostly offline
& Better learner-relevant preference alignment than offline preference optimization
& Limited by generated candidates, especially under sparse refresh or rare desirable behaviors \\

On-policy verifier / process guidance
& On-pol.
& Learner intermediate states
& Process feedback, verifiers
& Fine-grained learner-state reshaping
& Weakens when the same verifier mainly curates offline data
& Denser local correction than terminal rewards alone
& Depends on verifier faithfulness and reachable-state coverage \\

Iterative refinement / self-improvement
& Mixed or on-policy-like
& Learner-generated candidates and refinement traces
& Critique, revision, reranking, or self-improvement signals
& Usually reshaping
& Weakens when optimization shifts to a fixed retained batch
& Improves behavior through repeated self-correction and refinement
& Weak critique may reinforce superficial patterns, and refinement gains may transfer poorly to the final policy\\

On-policy teacher-guided transfer
& On-pol.
& Student rollouts
& Teacher-guided targets on learner rollouts
& Usually reshaping; sometimes consolidation-dominant
& Weakens when cached traces dominate fresh correction
& State-aligned teacher transfer; amortization of online gains
& Teacher cost, weak correction, or superficial retention \\

Hybrid pipelines
& Mixed
& External + learner trajectories
& Multiple stage-specific interfaces
& Coordinated expansion, reshaping, consolidation
& Depends on stage order and which stage removes the main bottleneck
& Strong for multi-bottleneck systems
& Sensitive to transition calibration and stage-transition attrition \\

\bottomrule
\end{tabularx}
\end{table*}

\subsubsection{Borderline Cases and Cross-Cutting Methods}
\label{subsubsec:comparative_borderline_merged}

Modern post-training includes many cases that are hybrid, boundary-crossing, or otherwise difficult to place under a single family label. Table~\ref{tab:borderline_cases_unified_lens} collects representative examples in a deliberately compact stress test.

\begin{table*}[t]
\centering
\scriptsize
\setlength{\tabcolsep}{3.8pt}
\renewcommand{\arraystretch}{1.12}
\caption{Representative borderline cases under the unified lens. The point is not to force each method into a single immutable box, but to show how provenance, supervision interface, and dominant role can be separated when family labels alone become ambiguous.}
\label{tab:borderline_cases_unified_lens}
\begin{adjustbox}{max width=\textwidth}
\begin{tabular}{>{\raggedright\arraybackslash}p{3.0cm} >{\raggedright\arraybackslash}p{2.0cm} >{\raggedright\arraybackslash}p{2.0cm} >{\raggedright\arraybackslash}p{2.6cm} >{\raggedright\arraybackslash}p{4.8cm}}
\toprule
\textbf{Method pattern} & \textbf{Provenance} & \textbf{Supervision interface} & \textbf{Typical dominant role} & \textbf{Why the assignment is non-trivial} \\
\midrule
Offline preference optimization vs. online preference optimization
& Offline: off-policy; online refresh: on-policy-like
& Pairwise or ranked preferences
& Usually reshaping
& The preference objective family stays similar while trajectory provenance changes with whether compared candidates are refreshed from the current learner \\

Verifier-guided offline filtering vs. verifier-guided on-policy correction
& Offline filtering: off-policy; rollout correction: on-policy
& Verifier scores, process checks, reranking
& Offline: usually reshaping, sometimes selective expansion; online: learner-state reshaping
& The same verifier can curate a static dataset or directly supervise learner-generated states; the evaluator does not by itself determine the regime \\

Teacher-generated traces vs. teacher correction on learner rollouts
& Teacher traces: off-policy; teacher correction: on-policy
& Teacher-guided targets
& Teacher traces: expansion or consolidation; learner correction: reshaping or consolidation
& The teacher mechanism is shared, but what changes is where the student is actually optimized \\

Self-improvement with cached trajectories
& Mixed or staged
& Critique, revision, verification, replay
& Usually reshaping, with possible later consolidation
& Fresh learner-state correction gradually mixes with stored traces, softening the provenance boundary and making the pipeline neither purely online nor purely offline \\

Search-heavy induction followed by policy transfer
& Mixed across stages
& Search/verifier signals followed by distillation or policy transfer
& Expansion first, then consolidation
& Strong behavior may first appear only under expensive search, so the key scientific question becomes whether later transfer preserves the induced behavior once the scaffold is removed \\
\bottomrule
\end{tabular}
\end{adjustbox}
\end{table*}

These cases show why provenance, supervision interface, and functional role should be analyzed separately, especially in mixed or staged systems where local correction, replay, and teacher guidance may coexist without collapsing into a single label. Autoregressive models add one further subtlety: local action improvements may have downstream state consequences, so a method that appears to be a local correction at one level may still induce broader trajectory-level support expansion through recursive rollout effects.

\subsection{What This Comparative Lens Does and Does Not Claim}
\label{subsec:hybrid_comparative_scope_limits}

This comparative lens is explanatory rather than exhaustive. Its purpose is to clarify how major post-training paradigms differ in trajectory provenance, supervision interface, and dominant functional role, not to assign immutable labels to method families or to imply that empirical outcomes such as capability gain, robustness, generalization, or forgetting can be read off directly from those labels alone. Its value lies in providing a structured language for interpreting mixed and multi-stage systems, especially when different bottlenecks dominate at different stages.

\subsection{Section Takeaways}
\label{subsec:hybrid_comparative_takeaways}

Hybrid pipelines are best understood as coordinated multi-stage compositions rather than as a third learning regime. Their scientific importance lies not only in how they sequence bottlenecks, but also in how they manage stage transition, preservation, and attrition across changing supervision regimes. Under the unified lens, their main value is therefore comparative as much as procedural: they reveal why strong systems often combine rather than replace major post-training paradigms.

%% file: sections/emerging_directions_and_open_problems.tex
\section{Emerging Directions and Open Problems}
\label{sec:emerging_open}

The field is already moving in several recognizable directions, but key scientific and evaluative questions remain unresolved. This section therefore separates \emph{emerging directions}---places where the literature is already moving in a visible way---from \emph{open problems} for which current evidence, measurement, or theory remains insufficient. The goal is not to predict a single dominant recipe, but to clarify which questions are becoming central as post-training shifts from isolated objective design toward coordinated behavioral system design.

\subsection{Emerging Directions}
\label{subsec:emerging_directions}

These trends do not point toward a single new objective; rather, they suggest a shift toward more policy-aware supervision, stronger guided correction on learner-generated states, tighter coupling between off-policy and on-policy updates, more explicit cross-stage preservation, and increasingly adaptive multi-stage system design.

\subsubsection{From Vanilla SFT to Model-Aware and Distribution-Aware Supervision}
\label{subsubsec:emerging_policy_aware_sft}

A first emerging direction is the growing redesign of supervised fine-tuning itself. Rather than treating SFT as fixed next-token imitation on externally supplied demonstrations, recent work increasingly adapts the objective, regularization, or training data to the learner's current policy, capability, or pretrained distribution \citep{wu2025dft, zhu2025psft, li2025beyond, zhu2026anchored, yang2024self, zhao2025mindgap, zhang2025datafit, gupta2025selective}. One line of work modifies the SFT objective directly, for example by reweighting token updates, constraining policy drift, or choosing probability-based objectives that better match the model's capability regime. Another line of work changes the supervision distribution itself, for example by selecting responses that better fit the target model, reusing the model's own correct outputs, or rewriting trajectories to shrink the policy gap before optimization.

The important shift is not that these methods leave the off-policy regime. It is that off-policy supervision is increasingly being designed with the learner's actual distribution in mind. Their shared goal is to make externally supplied supervision less mismatched to the learner's current distribution, less destructive to previously effective behavior, and more targeted to the actual bottleneck---whether support expansion, behavior reshaping, or attrition control. In this sense, the field is moving away from vanilla likelihood maximization over fixed demonstrations and toward a more model-aware, distribution-aware form of off-policy post-training.

\subsubsection{From Unguided Rollouts to Guided On-Policy Learning}
\label{subsubsec:emerging_guided_onpolicy}

A second major direction is the shift from unguided rollout optimization toward more guided forms of on-policy learning. As reasoning-intensive and agentic workloads become more important, the limits of blind rollout plus sparse terminal reward have become more visible. Models increasingly fail not because they never saw the right final answer, but because they do not reliably enter, sustain, or recover along useful reasoning paths under their own policy. This has driven growing interest in methods that guide on-policy learning more explicitly through rollout-time hints, self-explanations, templates, structured prompting, or learned self-correction behaviors that help the learner enter and sustain more useful reasoning paths \citep{liao2026sage, expo2025, wu2025templaterl, hipo2026, zhou2025ScaffoldedRL}.

These methods do not leave the on-policy regime. What changes is how learner-generated rollouts are sampled, guided, and corrected. Their shared goal is to make learner-state optimization less reliant on blind exploration, less vulnerable to sparse or collapsed learning signals, and more capable of reaching useful regions that the unaided policy would struggle to discover reliably. The dominant role often remains policy reshaping on learner-generated states, but increasingly with structured rollout-time guidance that can also support selective expansion of what becomes effectively reachable under the learner's own rollouts.

\subsubsection{From Rigid Stage Separation to Interleaved or Unified SFT--RL Training}
\label{subsubsec:emerging_unified_sftrl}

A third direction is the move from rigid stage separation toward interleaved or unified SFT--RL training. Increasing evidence suggests that a hard SFT $\rightarrow$ RL transition can be brittle: SFT and RL compensate for different weaknesses, and naively sequencing them can create cold-start difficulty, imitation--exploration imbalance, unstable stage transitions, or gradient interference. Recent work therefore explores more continuous ways of coupling external supervision with learner-driven optimization \citep{ma2026interleave, yan2025luffy, fu2025srft, liu2025uft, zhang2025chord, wang2025grao}.

These methods do not erase the distinction between off-policy and on-policy learning, but they do make the interaction between them more continuous. Their common goal is to soften the handoff between support-providing off-policy supervision and learner-state reshaping, so that useful external guidance remains available when purely on-policy optimization is still insufficient, without collapsing training into rigid imitation. The field is therefore moving away from treating SFT and RL as cleanly separable phases and toward treating their coupling itself as a design problem.

\subsubsection{Cross-Stage Preservation, Specialist Synthesis, and Distillation}
\label{subsubsec:emerging_cross_stage_preservation}

A fourth emerging direction is the increasing centrality of cross-stage preservation as a systems design concern. Strong behavior often first appears in a costly, scaffold-dependent, or domain-specialized form, after which later stages attempt to preserve, merge, or compress it into a cheaper and more deployable model \citep{gu2024minillm, zhang2025distill_rewards, sang2026crisp, luo2026agentark, dixit2026aletheia, liu2025deepseek, glm5team2026glm5}. In recent large-system recipes, this increasingly takes the form of specialist synthesis or cross-stage distillation, where behaviors acquired in earlier stages or specialized experts are distilled forward rather than left vulnerable to later stage transitions.

The central issue is no longer only how useful behavior is first induced. It is whether that behavior survives transfer, merging, compression, and handoff across stages, domains, and deployment regimes. This trend therefore reflects the growing importance of behavioral consolidation as a systems design objective rather than a late-stage implementation detail.

\subsubsection{From Offline Compression to On-Policy Distillation}
\label{subsubsec:emerging_onpolicy_distillation}

A fifth, more specific direction within this broader preservation trend is the growing prominence of on-policy distillation, with self-distillation becoming an especially important variant. Rather than treating distillation mainly as an offline, post hoc stage, recent work increasingly applies teacher guidance on the learner's own rollout states or carries improvements forward through distillation during continuing optimization \citep{agarwal2024opd, gu2024minillm, ye2026opcd, zhao2026selfDistilledReasoner}. This makes transfer and preservation more tightly coupled to the learner's visited states and may help retain improvements that would otherwise remain fragile under later rollout. A practical attraction of this direction is that it combines the learner-state relevance of on-policy training with denser teacher-derived supervision than sparse end-of-trajectory reward alone.

These methods remain on-policy even when they serve transfer or preservation functions. Their distinctive emphasis is to make consolidation directly responsive to learner-generated states and rollout-dependent improvements. For this reason, they lie near the boundary between learner-state reshaping and behavioral consolidation, showing how preservation can be tied more closely to on-policy learning.

\subsubsection{From Isolated Objectives to Adaptive Multi-Stage Systems}
\label{subsubsec:emerging_adaptive_agentic}

At a broader systems level, a sixth emerging direction is the shift from isolated objective comparisons toward adaptive multi-stage system design. Recent public system reports increasingly describe staged combinations of instruction tuning, preference optimization, online RL, tool-use adaptation, and later transfer or deployment-oriented consolidation rather than single-objective pipelines \citep{grattafiori2024llama, liu2025deepseek, glm5team2026glm5, yang2026nemotron, xiao2026mimo}. In this sense, post-training is increasingly becoming a systems problem of coordinating regimes, supervision interfaces, and preservation mechanisms under realistic constraints.

This shift does not make trajectory provenance, supervision interface, or functional role less important; it makes their composition across stages more important. One extension of this trend is the growing importance of agentic post-training, where models must act over longer horizons, interact with tools or environments, and remain robust under multi-step execution. More broadly, future post-training systems are likely to depend less on any single dominant objective and more on adaptive coordination across stages, signals, and deployment settings.

\subsection{Open Problems}
\label{subsec:open_problems}

The main unresolved questions no longer concern only whether one more objective or algorithm can improve benchmark scores. They increasingly concern how to diagnose behavioral bottlenecks, choose regimes and supervision interfaces accordingly, determine what behavior is transferable across stages, measure preservation and forgetting, evaluate pipeline-based systems, and explain the resulting empirical patterns. For clarity, the open problems below are organized into six higher-level clusters, summarized in Table~\ref{tab:open_problems_map}.

\begin{table*}[t]
\centering
\scriptsize
\setlength{\tabcolsep}{4.2pt}
\renewcommand{\arraystretch}{1.12}
\caption{Open-problems map under the unified post-training framework. The rows summarize six concrete problem clusters spanning support diagnosis, interface choice, transferability, preservation, system evaluation, and mechanistic explanation. Together, they reflect the broader shift from isolated objective design toward multi-stage behavioral system design.}
\label{tab:open_problems_map}
\begin{adjustbox}{max width=\textwidth}
\begin{tabular}{>{\raggedright\arraybackslash}p{3.3cm} >{\raggedright\arraybackslash}p{2.3cm} >{\raggedright\arraybackslash}p{4.8cm} >{\raggedright\arraybackslash}p{4.0cm}}
\toprule
\textbf{Open problem} & \textbf{Main level} & \textbf{Core question} & \textbf{Why it matters} \\
\midrule

How should effective support and bottlenecks be diagnosed?
& Provenance + functional role
& How should effective support be defined and measured, and how can one tell whether failures arise from missing support, poor policy reshaping on learner-visited states, or both?
& Without this, regime choice and role-based claims remain heuristic and difficult to compare across methods \\

How should supervision interfaces be chosen?
& Supervision interface
& Given a particular bottleneck and trajectory provenance, when are token targets, preferences, rewards, verifiers, process feedback, or teacher-guided targets the most informative interface?
& Better interface choice could improve efficiency, reduce unnecessary online cost, and make stage design more principled \\

What exactly is transferable across stages?
& Transfer object + consolidation
& What is the relevant object of transfer---final answers, reasoning traces, local policies, search behavior, tool-use patterns, or something else---and which kinds of behavior survive transfer reliably?
& This is central to distillation, amortization, multi-stage compression, and realistic deployment \\

How should preservation, forgetting, and handoff fidelity be measured?
& Consolidation + system stability
& How can one quantify retention, portability, selective forgetting, and handoff fidelity once a target behavior has been specified?
& End-task scores alone do not reveal whether expensive behavior has actually been stabilized or whether prior competence has been silently lost \\

How should multi-stage systems be evaluated?
& System design
& How should one report stage-handoff quality, scaffolded versus amortized performance, and cost--performance tradeoffs in pipeline-based systems?
& Modern post-training is increasingly pipeline-driven rather than single-objective, so evaluation must move beyond benchmark deltas \\

What mechanisms explain expansion, reshaping, and consolidation?
& Cross-cutting scientific understanding
& Why do different interventions differ in support expansion, learner-state reshaping, consolidation success, forgetting, and transfer behavior?
& The field increasingly needs explanatory understanding, not only stronger recipes \\

\bottomrule
\end{tabular}
\end{adjustbox}
\end{table*}

\subsubsection{How Should Effective Support and Bottlenecks Be Diagnosed?}
\label{subsubsec:open_support_bottlenecks}

One of the most central unresolved issues in post-training is how to diagnose the learner's actual bottleneck. In practice, it remains difficult to determine whether a system fails because it never reliably reaches useful behavior, because it reaches relevant prefixes but makes poor local decisions within them, or because both problems coexist.

A central open problem is therefore how effective support should be defined and measured in empirically usable terms. The immediate goal is operational diagnosis rather than a full mechanistic explanation: to determine whether the dominant bottleneck lies in missing support, poor learner-state reshaping, or both. This raises familiar but still unresolved questions. Should support expansion be characterized at the level of full trajectories, or decomposed into changes in prefix reachability and state-conditional action availability? How should one distinguish genuinely new behavior that has become effectively reachable from probability reweighting within already reachable regions?

This matters because diagnosis should guide intervention choice. If a model mainly lacks exposure to good trajectories, then off-policy support expansion may be the right response; if it already reaches relevant prefixes but behaves poorly on its own rollouts, then on-policy reshaping may be more appropriate. Yet many realistic systems exhibit both kinds of failure at once, and the dominant bottleneck may shift over training. Progress here will therefore likely require bottleneck-sensitive measurements, matched regime comparisons, and stage-order ablations that can reveal whether a gain came from newly reachable behavior, better ranking within already reachable regions, or some combination of the two.

One promising but still incomplete empirical direction is to use completion-level empirical support and large-sampling pass@k as practical proxies for bottleneck diagnosis. Recent work has begun to operationalize empirical support under finite sampling and to use large-k pass@k to probe whether RLVR genuinely expands capability beyond the base model or mainly improves selection within already reachable behavior \citep{wu2025support, yue2025doesrl}. These developments do not yet provide a complete measurement of the state-action notion of effective support used in this survey, but they do provide an important early form of operational grounding. Progress beyond this will likely require tighter links between such completion-level proxies and finer-grained measurements of prefix reachability, local action quality, and stage-specific behavioral change.

\subsubsection{How Should Supervision Interfaces Be Chosen?}
\label{subsubsec:open_interface_choice}

Another major open problem concerns supervision interfaces. The field now has access to token-level imitation, pairwise preferences, scalar rewards, verifier judgments, process-level feedback, critique--revision signals, and teacher-guided targets. Yet there is still limited systematic understanding of which interfaces are most informative for which bottlenecks once the relevant trajectories have been exposed.

Part of the difficulty is that supervision interface is neither a top-level regime nor a trivial implementation detail. Given a particular trajectory provenance, interface choice determines how corrective information is actually attached to the relevant trajectories. This creates several unresolved questions. When is token supervision sufficient, and when is preference supervision more informative? When do scalar rewards become too weak compared with verifier-based or process-level feedback? When should teacher-guided targets supplement or replace evaluative supervision? And can different interfaces be composed adaptively across stages rather than fixed in advance?

A more principled science of interfaces should therefore ask not only which interface performs better in aggregate, but which interface is most informative for a particular behavioral problem once the relevant states are already available. A token-target update, a preference update, and a verifier-guided update may all operate on similar data, yet differ sharply in what distinctions they make salient and how efficiently they correct behavior.

In practice, progress here will likely require controlled comparisons that hold bottleneck type and trajectory provenance as fixed as possible while varying only the supervision interface. The crucial experimental principle is to compare token targets, preferences, rewards, verifiers, process feedback, and teacher-guided targets on the same underlying behavioral problem rather than across loosely matched settings.

\subsubsection{What Exactly Is Transferable Across Stages?}
\label{subsubsec:open_transferability}

A further open problem concerns what exactly is being transferred or consolidated across stages. Some pipelines appear to preserve final-answer quality; others aim to carry forward reasoning traces, local action tendencies, search outcomes, critique--refinement behavior, tool-use patterns, or some mixture of these. Yet the relevant behavioral object is often left implicit, even though different objects may behave very differently under transfer.

This raises a deeper question: which kinds of behavior are actually transferable, and which are not? A student may copy the final format of a strong teacher while failing to inherit the teacher's intermediate reasoning behavior. A policy may preserve benchmark reward while losing robustness under different prompts or rollout conditions. Search- or verifier-heavy behaviors may prove difficult to amortize even when final outputs remain superficially similar. More broadly, some induced behaviors appear easy to distill or compress, whereas others collapse once teacher support, verifier scaffolding, search, or multi-sample exploration is removed.

These are not peripheral implementation details. They bear directly on what behavioral consolidation should mean in the first place. If the transferred object is left underspecified, then claims about distillation, amortization, or multi-stage preservation remain hard to interpret. The relevant scientific question is therefore not merely whether transfer ``works,'' but what is being transferred, which behavioral categories are systematically more or less transferable, and why.

Progress here will likely require more explicit transfer ontologies and evaluation protocols that treat final answers, reasoning processes, local decision policies, search behavior, and tool-interaction patterns as potentially distinct objects of consolidation rather than as a single undifferentiated phenomenon.

\subsubsection{How Should Preservation, Forgetting, and Handoff Fidelity Be Measured?}
\label{subsubsec:open_preservation_measurement}

Closely related, but distinct, is the question of how preservation should be evaluated once the target behavioral object has been specified. Current reporting often emphasizes end-task performance after transfer, adaptation, or reconsolidation, but this gives only a partial view. If an expensive teacher, search process, or online RL stage discovers useful behavior, the scientific question is not only whether a later student scores well on a benchmark. It is whether the relevant behavior has actually been retained, stabilized, and made portable under deployment conditions that no longer include the original inducing machinery.

This makes preservation and forgetting a joint measurement problem. One needs to know not only whether useful behavior survives handoff, but also whether other previously reliable behavior is silently degraded in the process. Retention under stage transition, portability across prompt distributions, robustness after scaffold removal, and fidelity of intermediate reasoning or local decision patterns may all matter, depending on what the system is trying to preserve. At the same time, post-training rarely seeks perfect preservation of everything: later stages often need to update some behaviors aggressively while protecting others from attrition.

The open problem is therefore twofold. First, what metrics and evaluation protocols can capture retention, portability, handoff fidelity, and selective forgetting at the level of behavior rather than only final score? Second, how should systems balance adaptation against preservation when behavior is repeatedly moved across off-policy, on-policy, and transfer-based stages? Progress here will likely require consolidation-sensitive evaluation protocols, stage-handoff studies, and reporting conventions that go beyond end-task accuracy to measure whether expensive behavior has actually been stabilized, whether prior competence has been selectively preserved, and where support attrition has occurred during later adaptation.

\subsubsection{How Should Multi-Stage Systems Be Evaluated?}
\label{subsubsec:open_pipeline_evaluation}

As post-training becomes more explicitly pipeline-oriented, evaluation itself becomes a systems problem. Single-number benchmark improvements are no longer enough to describe systems that move through off-policy support expansion, on-policy policy reshaping, and later consolidation. A central open question is therefore how multi-stage post-training systems should be evaluated \emph{as systems} rather than only as collections of individual objectives.

Several issues follow immediately. What should count as a successful handoff between stages? How should one report whether useful behavior survives from one stage to the next? How should one distinguish scaffolded behavior that depends on expensive search, verifier interaction, or multi-sample reasoning from behavior that has actually been amortized into a cheaper deployment model? And how should cost--performance tradeoffs be compared when two systems achieve similar benchmark scores through very different intermediate machinery?

These are not merely reporting details. They are central to understanding whether post-training pipelines are scientifically principled or only superficially effective. If strong behavior appears only under large search budgets, privileged scaffolds, or costly intermediate procedures, then the key systems question is whether later stages preserve that behavior under realistic deployment conditions. This suggests the need for cost-aware reporting protocols, explicit stage-handoff metrics, and evaluations that compare scaffolded behavior with amortized deployment behavior rather than collapsing them into a single end score.

\subsubsection{What Mechanisms Explain Expansion, Reshaping, and Consolidation?}
\label{subsubsec:open_mechanisms}

A final cross-cutting open problem is mechanistic explanation. The field increasingly observes recurring empirical patterns: off-policy supervision often appears strong at importing useful behavior into practical reach; on-policy learning often appears stronger at correcting rollout-time failures on learner-generated states; distillation can preserve some induced behaviors but not others; and hybrid pipelines frequently outperform any single regime alone. Yet these regularities still lack a sufficiently precise explanatory account.

Three unresolved explanatory gaps seem especially important. First, when apparent gains are observed, what distinguishes genuine expansion of effectively reachable behavior from improved elicitation, ranking, or search support within regions that were already weakly available? In other words, when should post-training be understood as injecting capability and when should it be understood as better eliciting capability that was already latent in the model? This question is closely tied to capability boundary: without clearer mechanistic criteria, it remains difficult to say whether a method has actually enlarged what the model can reliably do or has instead made existing behavior easier to realize under a given prompting or rollout setup.

Second, why do different intervention regimes tend to succeed on different bottlenecks? Off-policy methods often appear strong when the learner lacks access to useful trajectories, while on-policy methods often appear stronger when the learner already reaches relevant states but behaves poorly within them. This suggests that differences in trajectory provenance may interact with rollout dynamics, state visitation, and local credit assignment in more systematic ways than current method-level narratives make explicit. A deeper account is still needed of why some interventions mainly broaden effective support, while others mainly reshape behavior inside already reachable regions.

Third, why are some induced behaviors readily consolidatable while others remain scaffold-dependent, brittle, or fragile under transfer? Distillation and related transfer stages can preserve certain gains across model handoff, compression, or scaffold removal, but they do not do so uniformly. This raises a broader mechanistic question about what makes a behavior portable across stages and deployment regimes, and what causes support attrition or transfer asymmetry when apparently useful behavior is moved from one training context to another.

These questions connect the survey's descriptive categories to the deeper scientific phenomena they are meant to explain. Without better mechanistic understanding, notions such as capability injection versus elicitation, capability boundary, catastrophic forgetting, support attrition, and transfer asymmetry remain suggestive but only partially resolved. The field therefore needs more work that links distributional objects, rollout dynamics, update structure, and stage handoff to the empirical differences observed among SFT, RL, distillation, and hybrid post-training.

Methodologically, this will likely require clearer operational objects, intervention-based comparisons, provenance swaps, interface-controlled studies, stage-order ablations, and reporting protocols that tie empirical gains to explicit explanatory claims rather than only benchmark deltas.

\subsubsection{Section Takeaways}
\label{subsubsec:open_problems_takeaways}

Taken together, these open problems show that the next phase of post-training research will depend less on proposing yet another isolated objective and more on building a sharper science of bottleneck diagnosis, interface choice, stage composition, transfer, and preservation across multi-stage systems.

%% file: sections/conclusion.tex
\section{Conclusion}
\label{sec:conclusion}

Post-training has become a central phase through which large language models are transformed from broadly pretrained sequence predictors into aligned, task-competent, and deployable systems. Yet the field is still often described through disconnected method names or objective families. This survey has argued that such labels remain useful, but are no longer sufficient as the primary organizing principle for understanding modern post-training.

The main perspective advanced here is to view LLM post-training as structured intervention on model behavior. Within that perspective, trajectory provenance provides the most informative top-level distinction: whether improvement is driven primarily by externally supplied trajectories or by learner-generated rollouts. Supervision interfaces then clarify how learning signals are attached, while the recurring roles of support expansion, policy reshaping, and behavioral consolidation help explain what different stages are mainly doing.

Seen in this way, modern post-training is better understood as a coordination problem than as a contest among isolated methods. Off-policy and on-policy learning are complementary regimes, and behavioral consolidation becomes central whenever useful induced behavior must survive handoff across stages, models, or deployment settings. Hybrid pipelines therefore matter not as a third regime, but as coordinated stage compositions whose success depends on whether each stage addresses a distinct bottleneck and whether the resulting behavior survives the next handoff.

More broadly, progress in post-training should be framed less as a search for a single best objective and more as a problem of behavioral systems design. Future advances will likely depend not only on stronger objectives, but also on better bottleneck diagnosis, better stage composition, and more faithful preservation of induced behavior across transitions.